\documentclass[acmtog,screen,authorversion,balance=false]{acmart}
\usepackage{graphicx}
\usepackage{amsmath}
\usepackage{hyperref}
\usepackage{caption}
\usepackage{xcolor}
\usepackage{multicol,multirow}
\usepackage{pifont}
\usepackage{tablefootnote}
\newcommand{\figTeaser}{
\begin{teaserfigure}
\centering
\includegraphics[width=\linewidth]{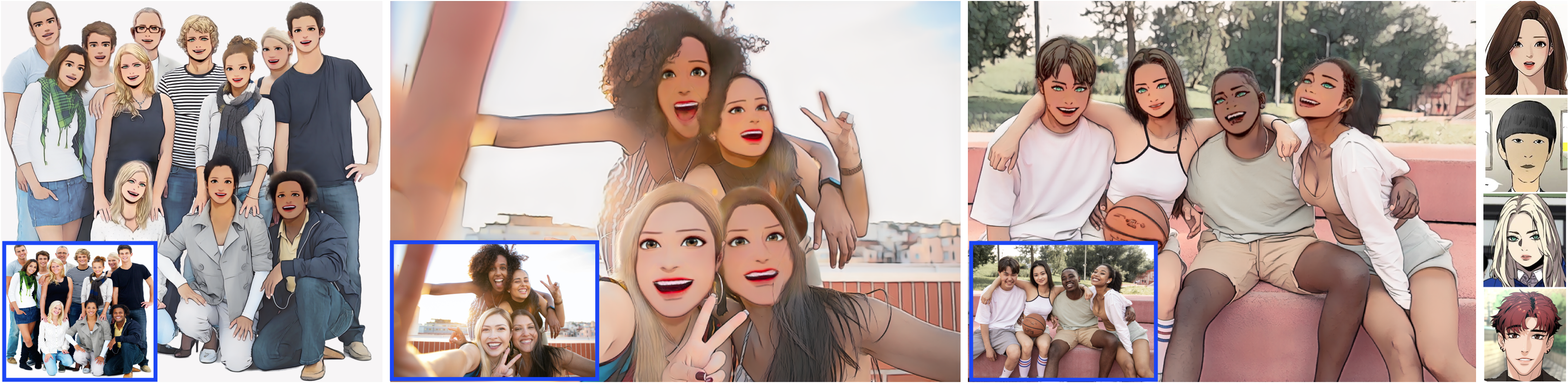}
\caption{Our system successfully stylizes full-body portraits while preserving the color diversity of the input photography. Webtoon characters (rightmost) are the representative examplar portraits of the target domain datasets. \textcopyright \textit{Obra Shalom Campo Grande, MS} \textcopyright \textit{Kampus Production} \textcopyright \textit{Monstera}}
\label{fig:teaser}
\end{teaserfigure}
}

\newcommand{\figFramework}{
\begin{figure}
\centering
\includegraphics[width=\linewidth]{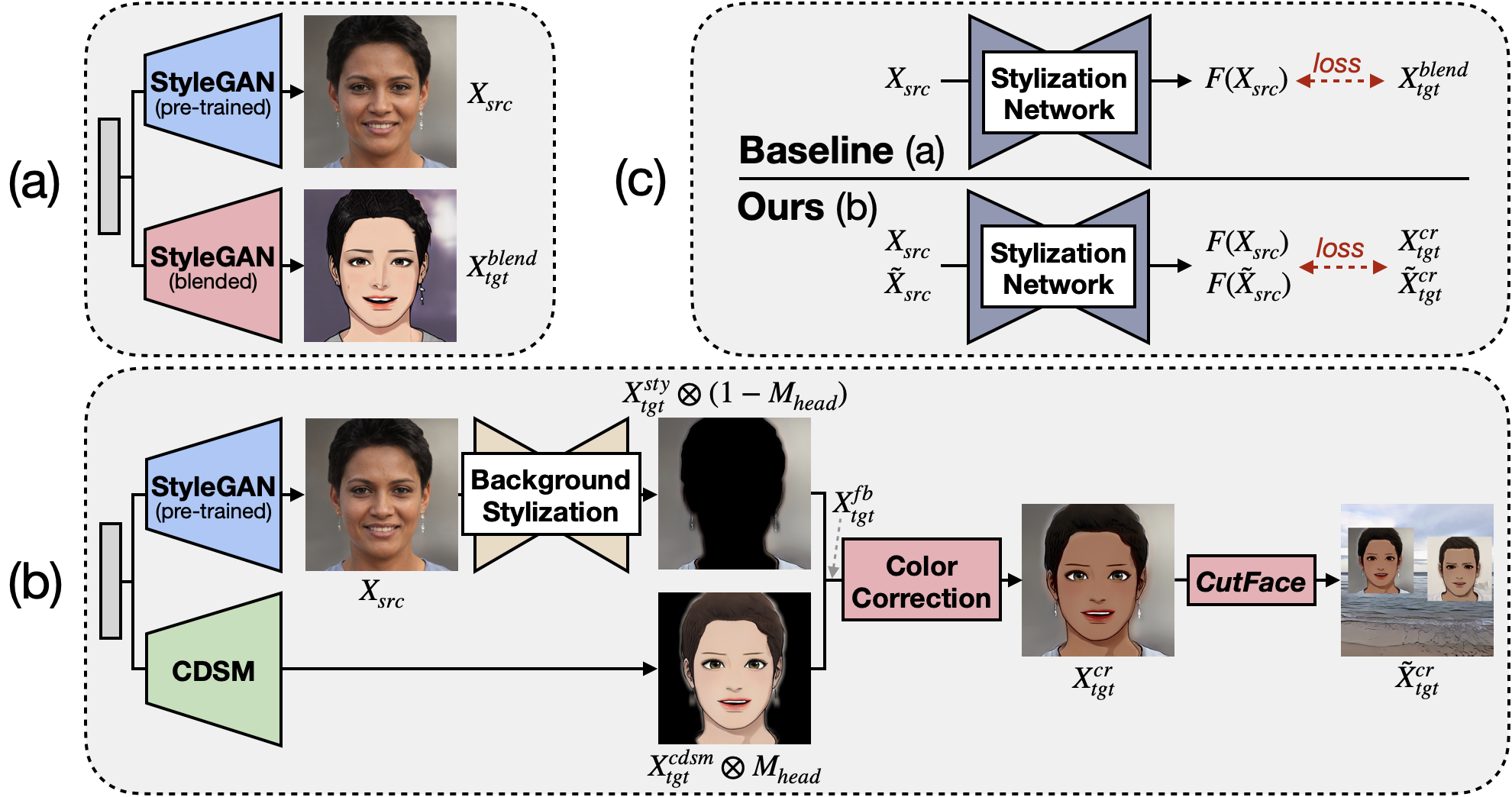}
\caption{\textbf{Overview.} \textbf{(a,c)} Two-stage approach [Spirin 2022]. It creates a synthetic-paired dataset with pre-trained and blended StyleGAN2 and then trains the stylization network using generated dataset.  \textbf{(b,c)} Upon this scheme, we introduce a novel data preparation stream to satisfy the practical demands that the vanilla two-stage approach cannot handle.}
\label{fig:framework}
\end{figure}
}

\newcommand{\figHistogram}{
\begin{figure}[t]
\centering
\newcommand{\h}{29.3mm}
\newcommand{\himg}{-2mm}
\centering
\makebox[\h][c]{\small(a) \textit{Head}: FFHQ}\hspace{\himg}
\makebox[\h][c]{\small(b) \textit{Head}: Baseline}\hspace{\himg}
\makebox[\h][c]{\small(c) \textit{Head}: Ours}\hfill
\\
\includegraphics[width=\h]{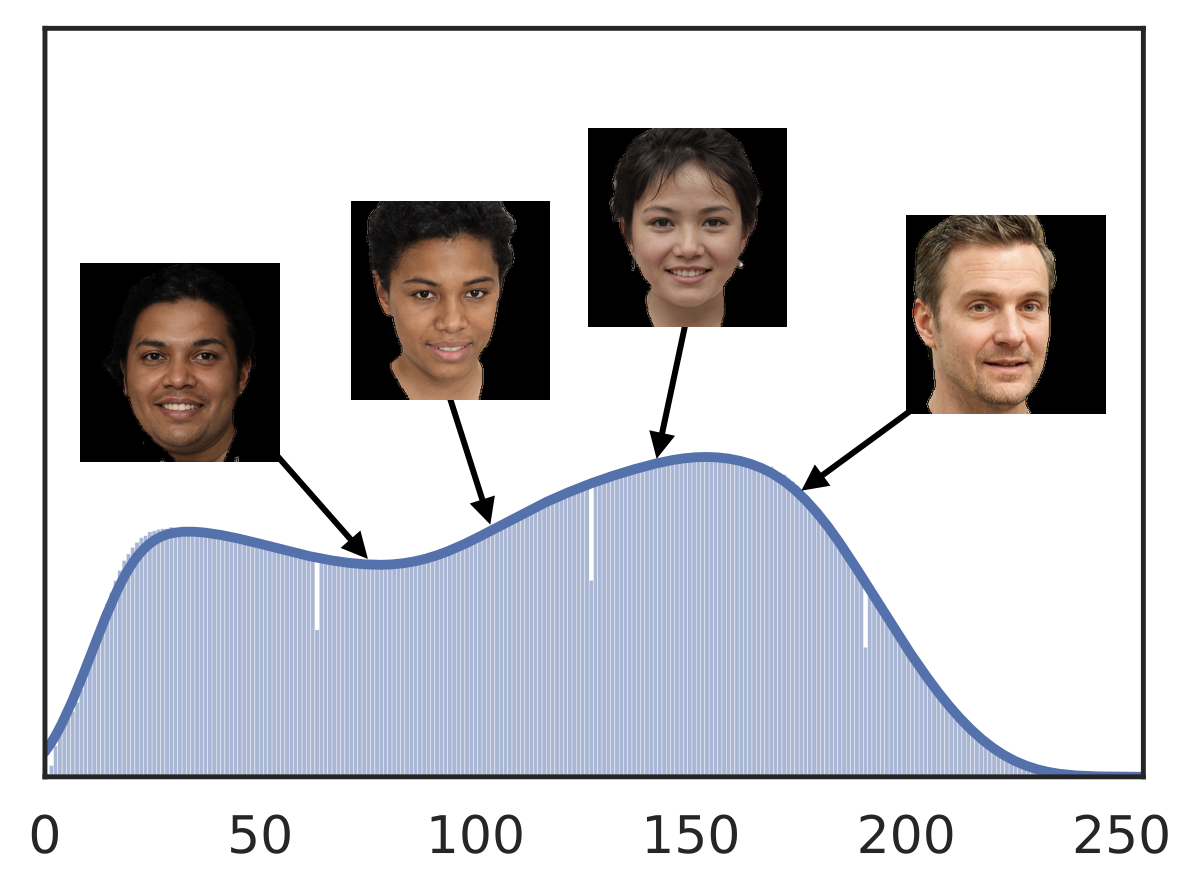}\hspace{\himg}
\includegraphics[width=\h]{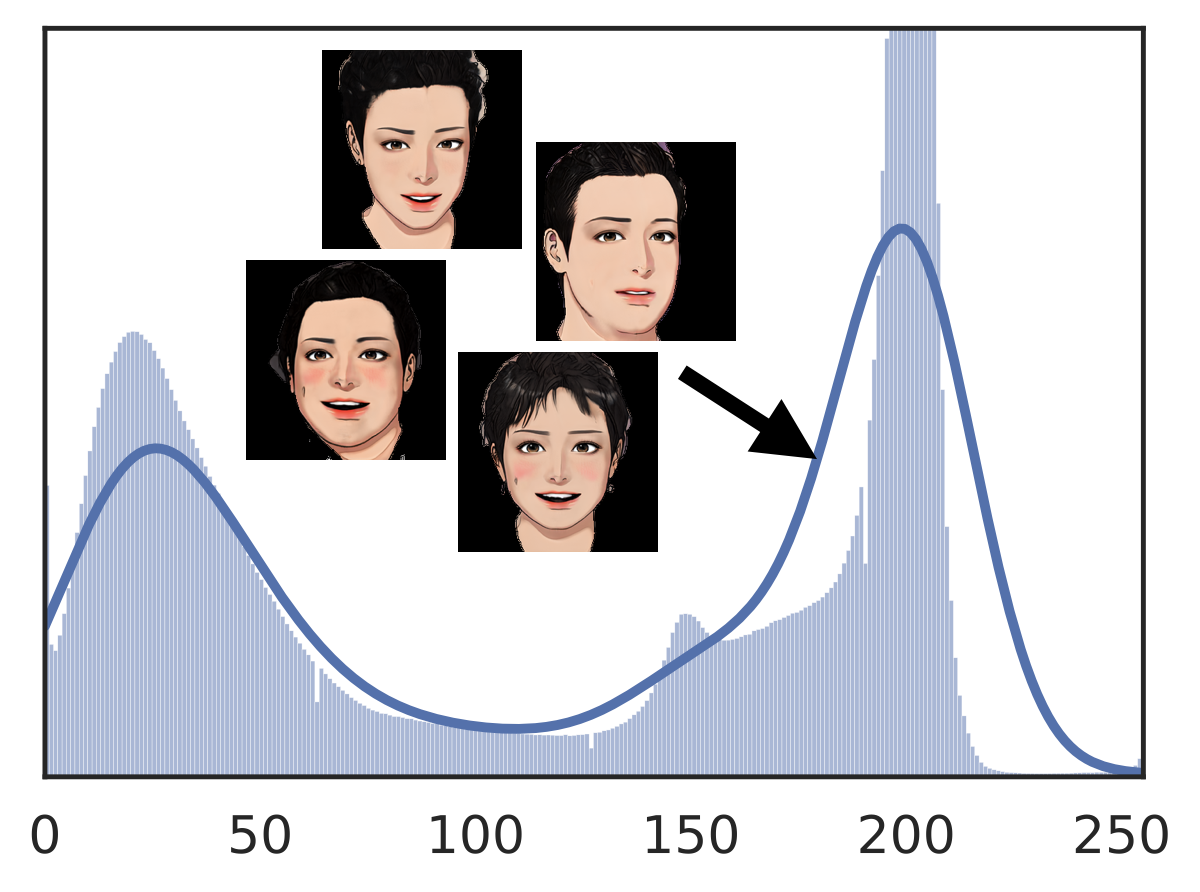}\hspace{\himg}
\includegraphics[width=\h]{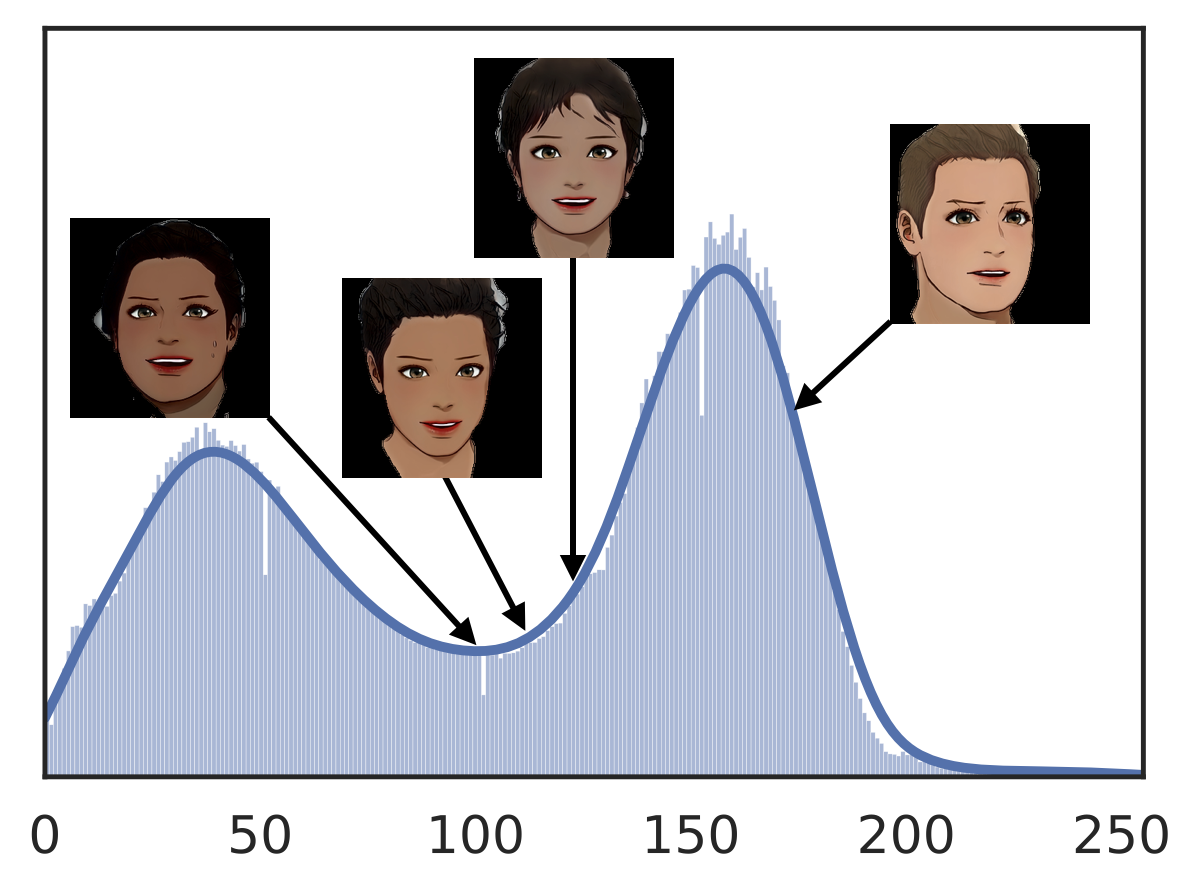}\hfill
\\\vspace{0.4em}
\makebox[\h][c]{\small(d) \textit{Background}: FFHQ}\hspace{\himg}
\makebox[\h][c]{\small(e) \textit{Background}: Baseline}\hspace{\himg}
\makebox[\h][c]{\small(f) \textit{Background}: Ours}\hfill
\\
\includegraphics[width=\h]{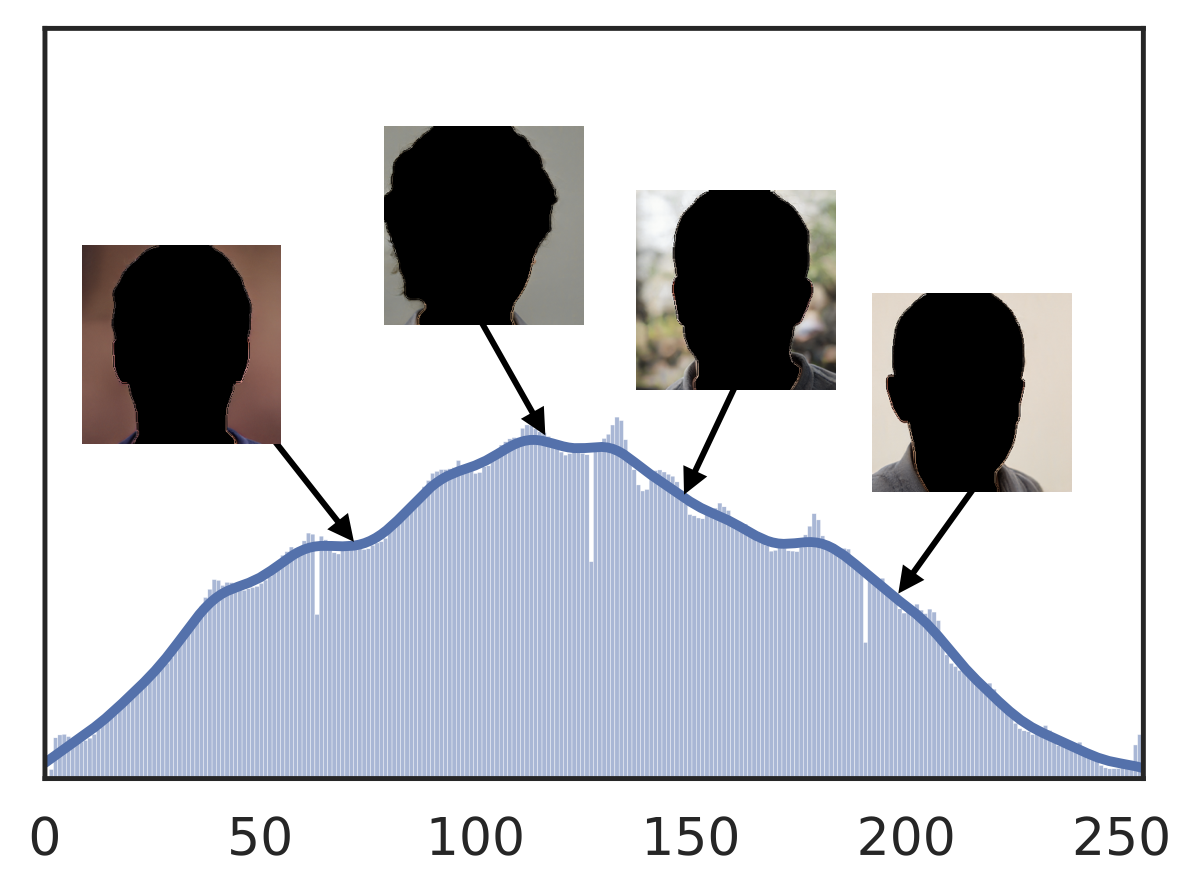}\hspace{\himg}
\includegraphics[width=\h]{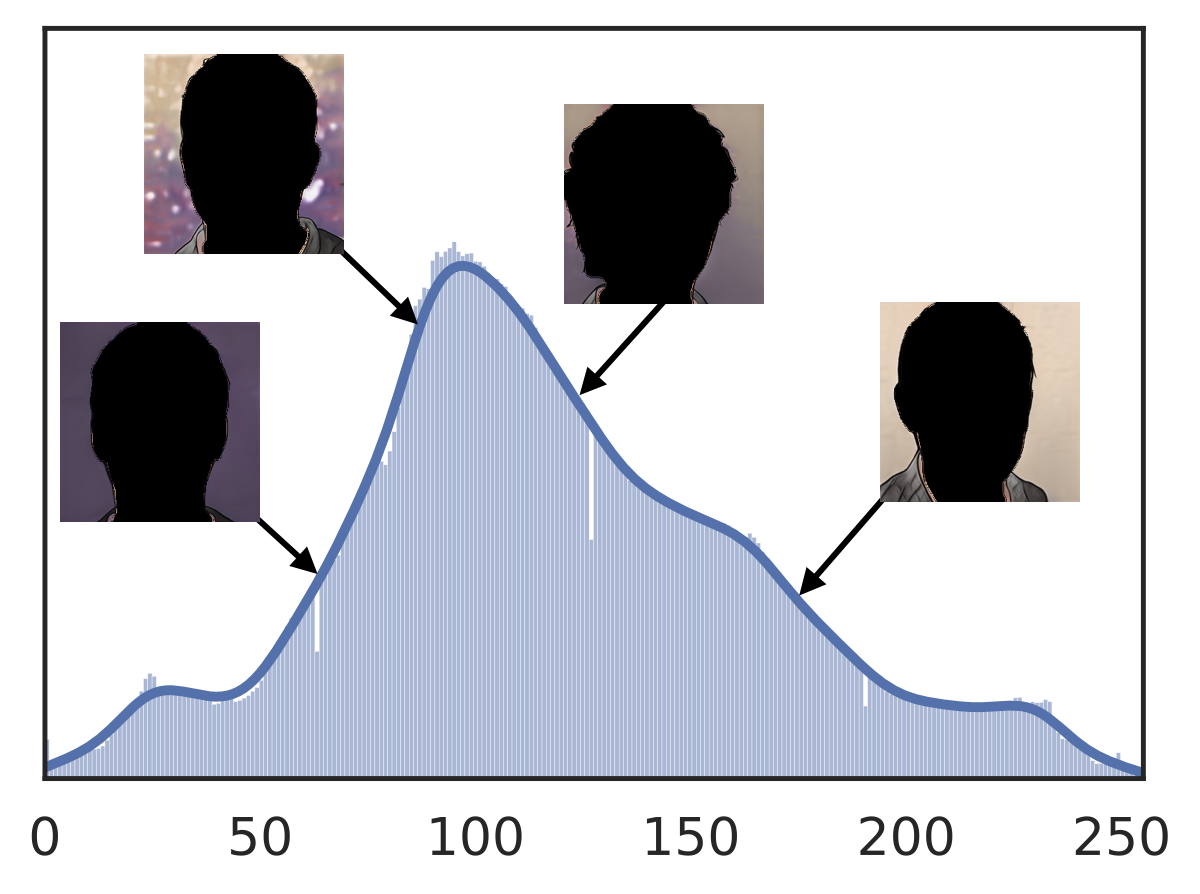}\hspace{\himg}
\includegraphics[width=\h]{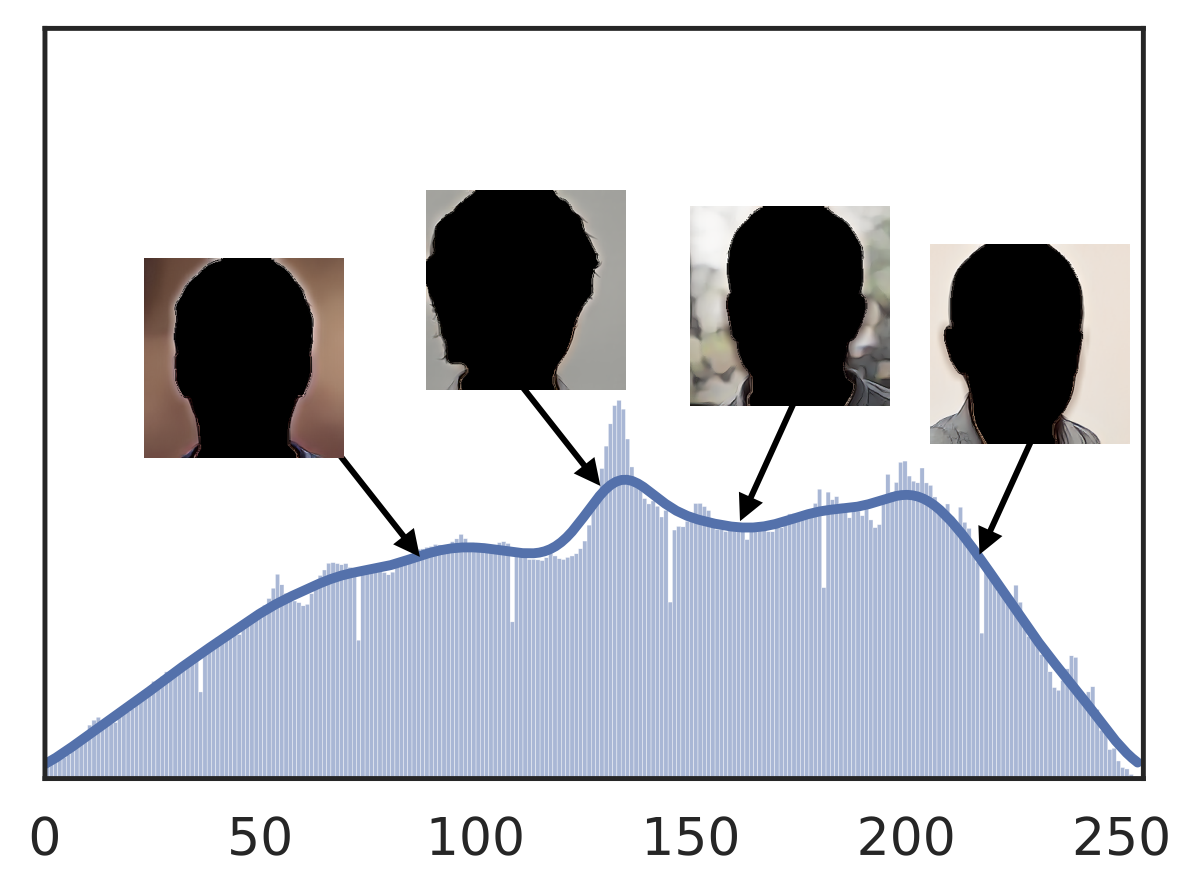}\hfill
\caption{\textbf{Pixel distribution ([0, 255]) of the paired dataset.} Note that FFHQ indicates the source domain. \textbf{(Top)} Pixel distribution of head regions. The baseline over-saturates the distribution toward a larger value, whereas our approach successfully restrains this. For all cases, a peak near 25 is caused by hair pixels. \textbf{(Bottom)} Pixel distribution of the background (outside of the head) regions. The baseline distorts the shape of the FFHQ distribution, whereas ours maintains the similar distribution shape.}
\label{fig:histogram}
\end{figure}
}

\newcommand{\figAblation}{
\begin{figure}[t]
\newcommand{\h}{20.6mm}
\newcommand{\himg}{0.01mm}
\newcommand{\hh}{27.7mm}
\newcommand{\hhimg}{0.01mm}
\centering
\makebox[\h][c]{\small(a) Source}\hspace{\himg}
\makebox[\h][c]{\small(b) Baseline}\hspace{\himg}
\makebox[\h][c]{\small(c) + FB-aware}\hspace{\himg}
\makebox[\h][c]{\small(d) + \textit{CutFace}}\hfill
\\\
\includegraphics[width=\h]{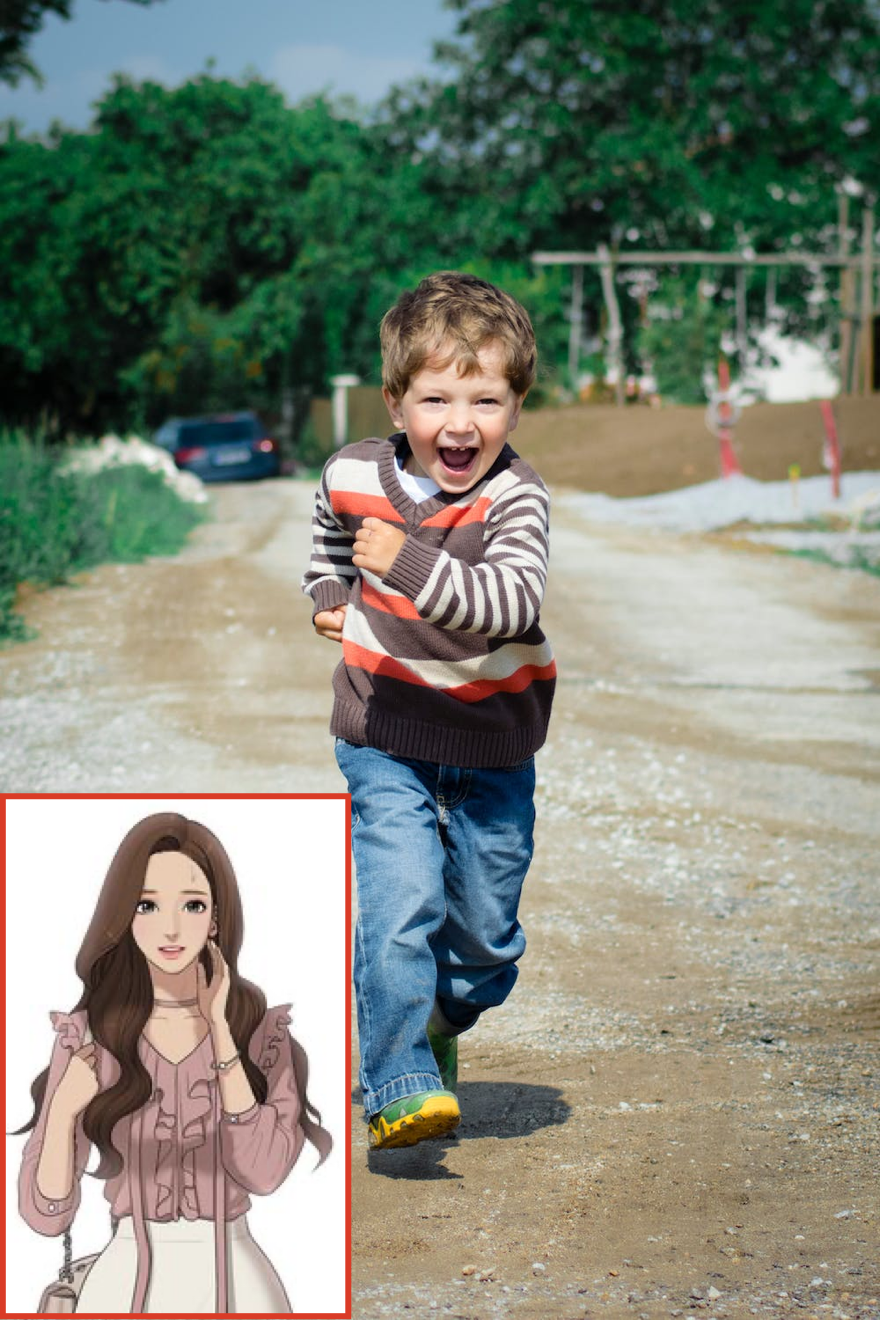}\hspace{\himg}
\includegraphics[width=\h]{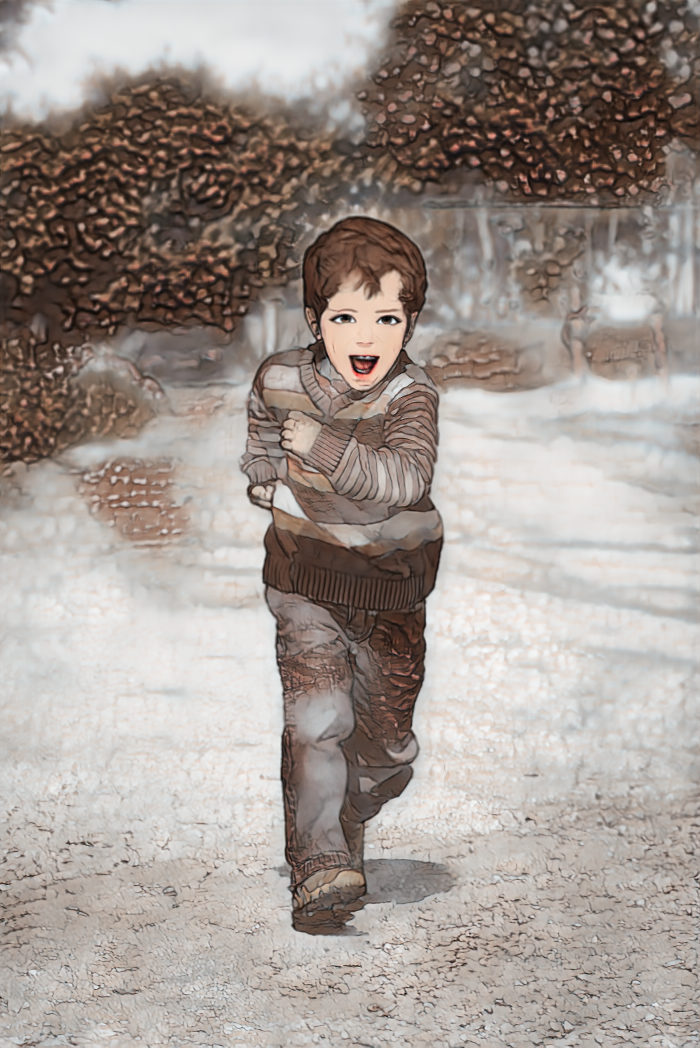}\hspace{\himg}
\includegraphics[width=\h]{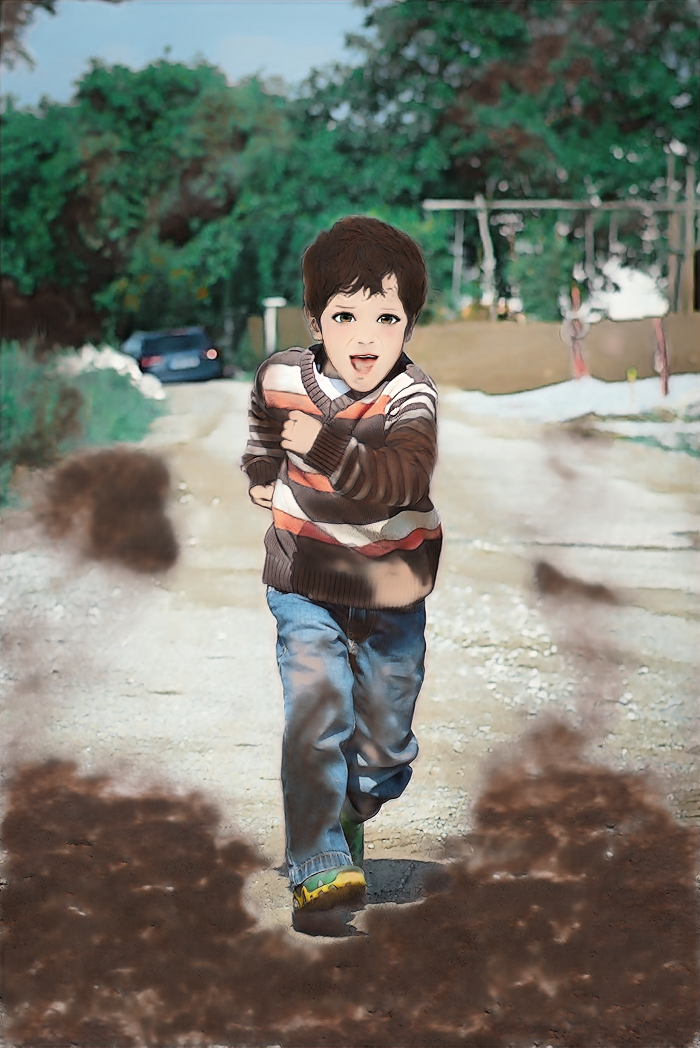}\hspace{\himg}
\includegraphics[width=\h]{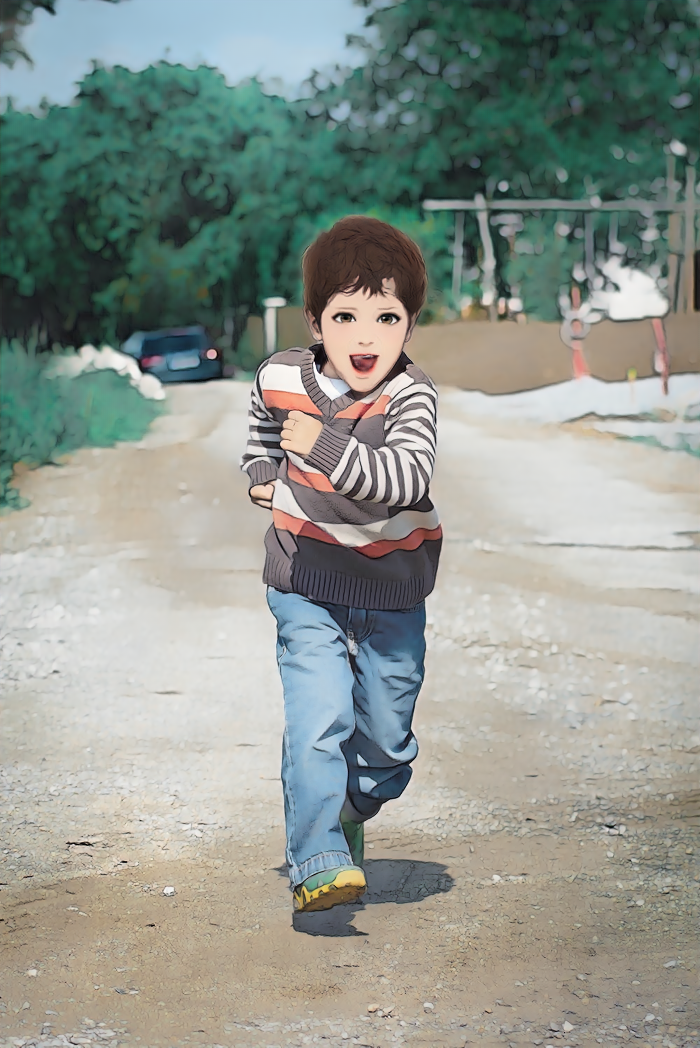}\hfill
\\\vspace{0.4em}
\makebox[\hh][c]{\small(e) Source}\hspace{\hhimg}
\makebox[\hh][c]{\small(f) w/o Color correction}\hspace{\hhimg}
\makebox[\hh][c]{\small(g) w/ Color correction}\hfill
\\
\includegraphics[width=\hh]{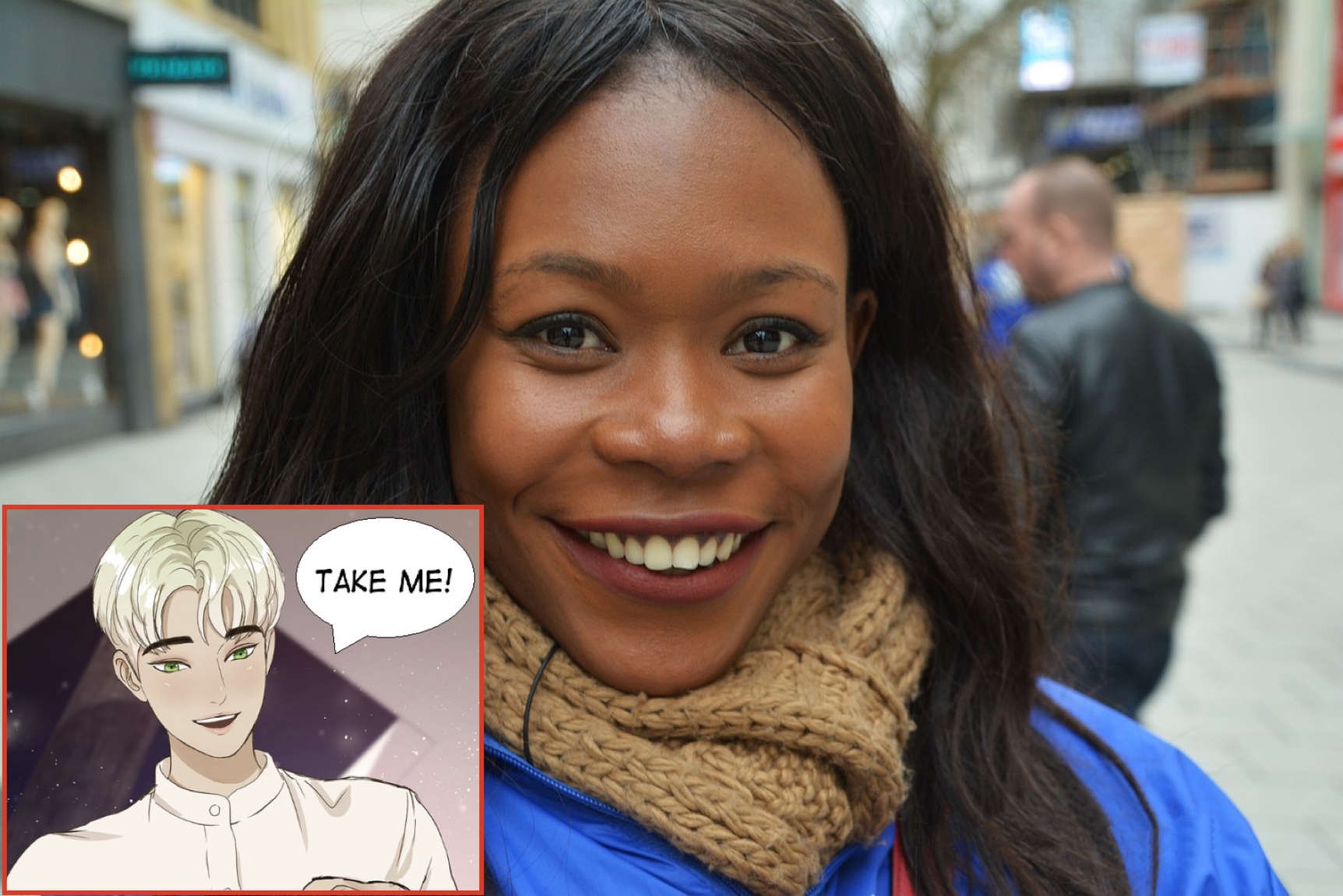}\hspace{\hhimg}
\includegraphics[width=\hh]{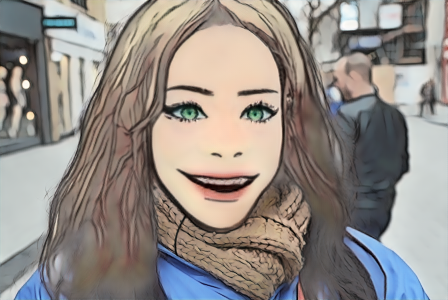}\hspace{\hhimg}
\includegraphics[width=\hh]{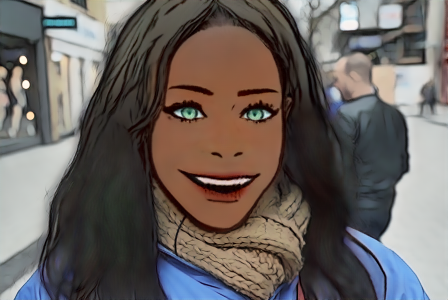}\hfill
\caption{\textbf{Ablation study.} Each proposed procedure successfully improves the quality, robustness, and ability to reflect skin color of the source photography. Baseline: \citet{arcanegan}. \textcopyright \textit{Luna Lovegood} \textcopyright \textit{terimakasih0}}
\label{fig:ablation}
\end{figure}
}

\newcommand{\figComp}{
\begin{figure*}[t]
\newcommand{\h}{43mm}
\newcommand{\himg}{1.0mm}
\centering
\makebox[\h][c]{\small(a) Source}\hspace{\himg}
\makebox[\h][c]{\small(b) AnimeGAN~\cite{chen2019animegan}}\hspace{\himg}
\makebox[\h][c]{\small(c) Two-stage~\cite{arcanegan}}\hspace{\himg}
\makebox[\h][c]{\small(d) WebtoonMe (ours)}\hfill
\\
\includegraphics[width=\h]{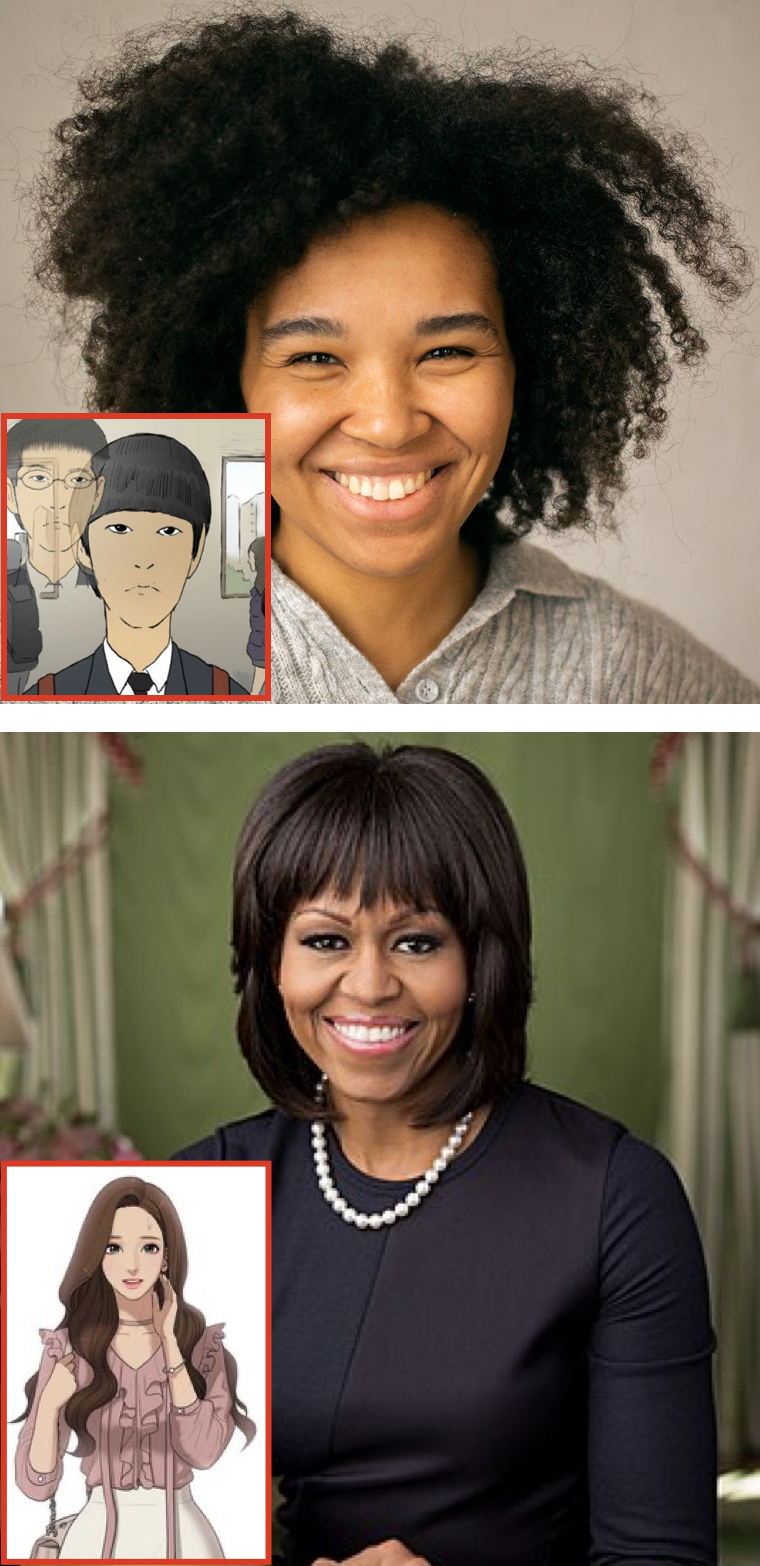}\hspace{\himg}
\includegraphics[width=\h]{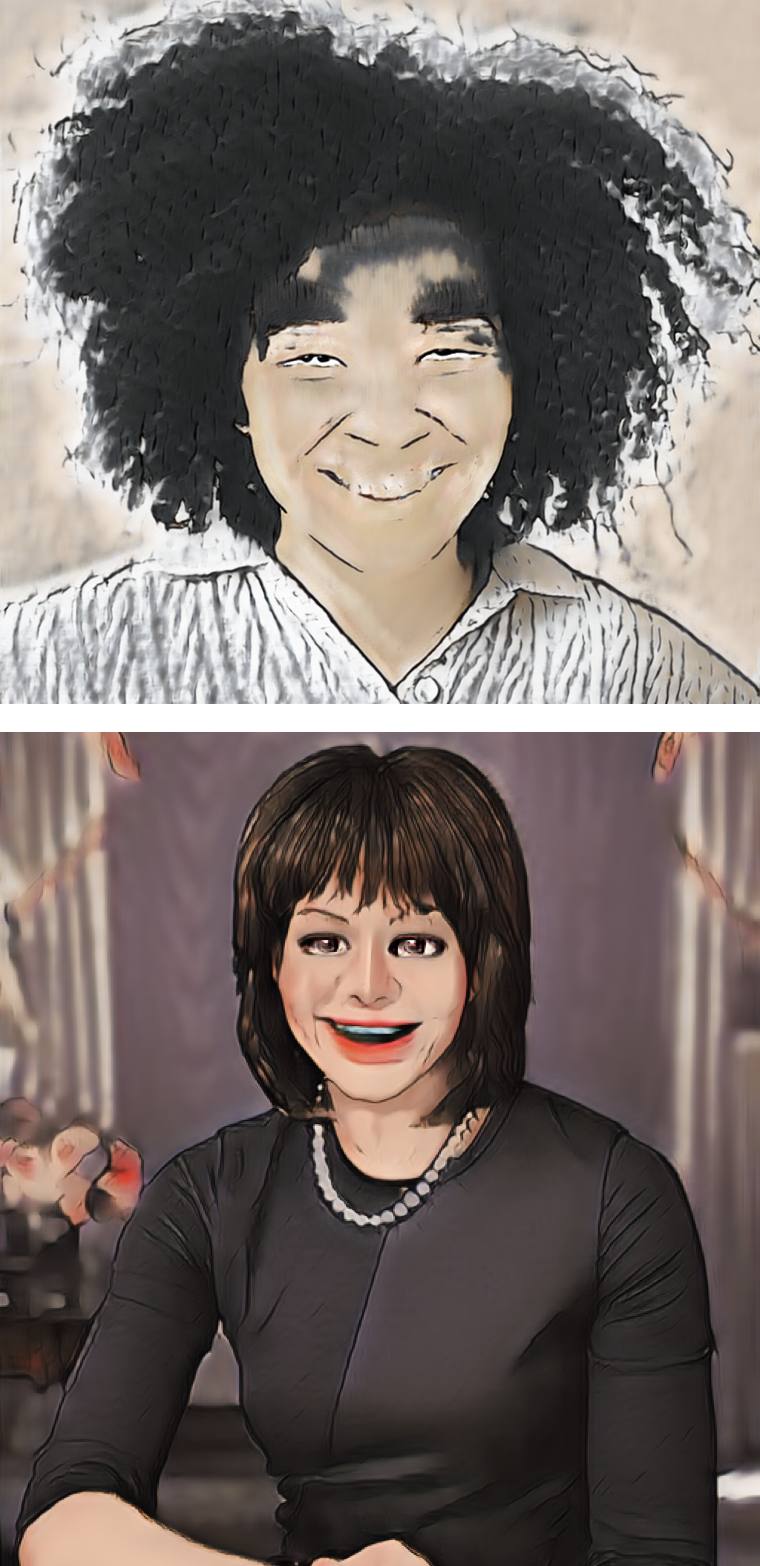}\hspace{\himg}
\includegraphics[width=\h]{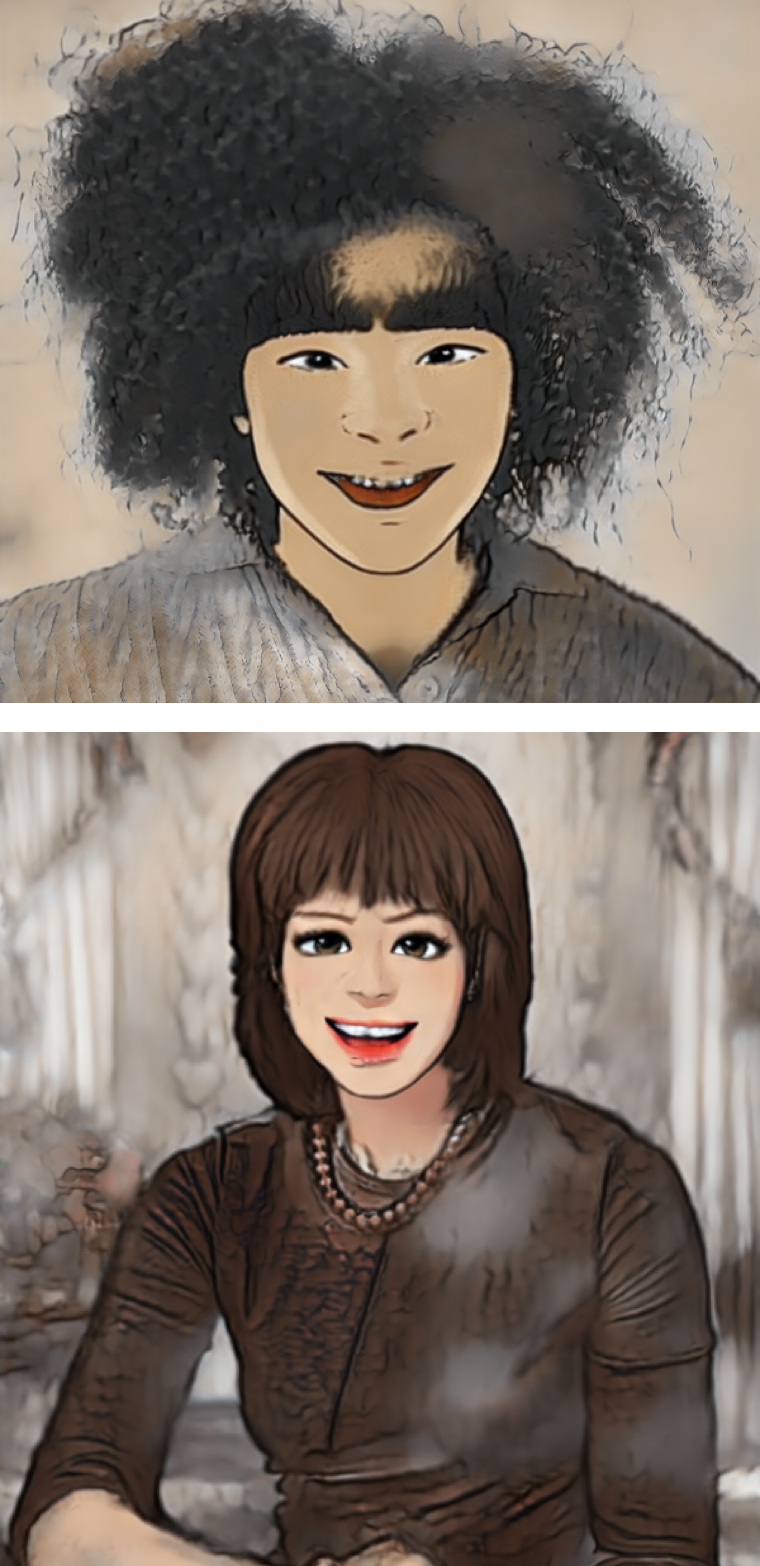}\hspace{\himg}
\includegraphics[width=\h]{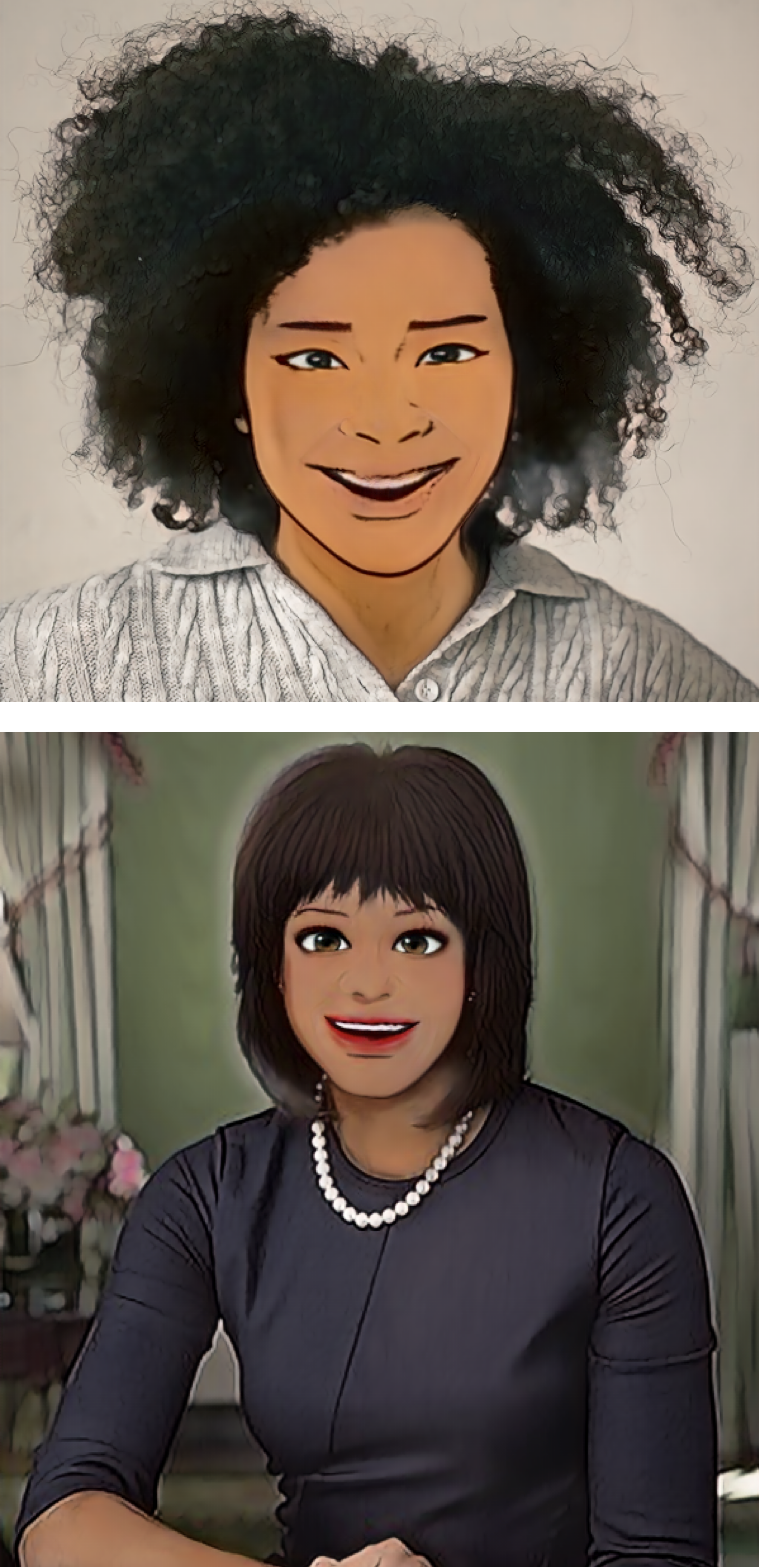}\hfill
\caption{\textbf{Qualitative comparison.} WebtoonMe shows superiority in both stylization quality and expression of color diversity. \textcopyright \textit{Monstera} \textcopyright \textit{White House}}
\label{fig:comp}
\end{figure*}
}

\newcommand{\tableContribution}{ 
\begin{table}[t]
\centering
\setlength\tabcolsep{4.2pt}
\caption{\textbf{Methodology comparison.} \textit{Image-to-image}: \citet{kim2019u}, \textit{Face stylization}: \citet{chong2021jojogan,kim2022cross}, \textit{Two-stage}: \citet{arcanegan}. This produces insufficient results in non-face regions.}
\begin{tabular}{l | c | c | c}
\hline
\multirow{2}{*}{Methodology} & \multicolumn{3}{c}{Practical requirements} \\
\cline{2-4}
& Full-body & Quality \& Robust. & Skin diversity \\
\hline\hline
Image-to-image & \xmark & \xmark & \xmark \\
Face stylization & \xmark & \cmark & \xmark \\
Two-stage & \cmark & \xmark & \xmark \\\hline\hline
Ours & \cmark & \cmark & \cmark \\
\hline
\end{tabular}
\label{table:contribution}
\end{table}
}

\newcommand{\tableComp}{ 
\begin{table}[t]
\centering
\setlength\tabcolsep{10pt}
\caption{\textbf{User study.} We compare our model with previous models that can stylize full-body portrait. AnimeGAN was trained on a synthetic paired dataset (identical to the two-stage method) owing to low data volume.}
\begin{tabular}{c | c | c | c}
\hline
\multirow{2}{*}{Method} & Quality & Robustness & Input Color\\
& (Face) & (Background) & Correction\\
\hline\hline
AnimeGAN & 1.4\% & 2.1\% & 1.4\% \\
Two-stage & 12.2\% & 2.1\% & 7.3\% \\
\hline\hline
Ours & \textbf{86.5\%} & \textbf{95.8\%} & \textbf{91.3\%} \\
\hline
\end{tabular}
\label{table:comparison}
\end{table}
}

\newcommand{\cmark}{\textcolor{teal}{\ding{51}}}
\newcommand{\xmark}{\textcolor{red}{\ding{55}}}
\newcommand{\Sref}[1]{Sec. \ref{#1}}

\newcommand{\Tref}[1]{Table \ref{#1}}
\newcommand{\Fref}[1]{Fig. \ref{#1}}

\AtBeginDocument{\providecommand\BibTeX{{
    \normalfont B\kern-0.5em{\scshape i\kern-0.25em b}\kern-0.8em\TeX}
}}

\copyrightyear{2022} 
\acmYear{2022} 
\setcopyright{acmcopyright}\acmConference[SA '22 Technical Communications]{SIGGRAPH Asia 2022 Technical Communications}{December 6--9, 2022}{Daegu, Republic of Korea}
\acmBooktitle{SIGGRAPH Asia 2022 Technical Communications (SA '22 Technical Communications), December 6--9, 2022, Daegu, Republic of Korea}
\acmPrice{15.00}
\acmDOI{10.1145/3550340.3564226}
\acmISBN{978-1-4503-9465-9/22/12}

\acmSubmissionID{124}

\begin{document}
\title{\mbox{WebtoonMe: A Data-Centric Approach for Full-Body Portrait Stylization}}
\mytitle{WebtoonMe: A Data-Centric Approach for Full-Body Portrait Stylization}

\author{Jihye Back}
\authornote{Both authors contributed equally.}
\email{1oojihye@webtoonscorp.com}
\author{Seungkwon Kim}
\authornotemark[1]
\email{mark.kim@webtoonscorp.com}
\author{Namhyuk Ahn}
\email{nhahn@webtoonscorp.com}
\affiliation{\institution{NAVER WEBTOON AI} \country{South Korea}}

\renewcommand{\shortauthors}{Jihye Back, Seungkwon Kim, Namhyuk Ahn}

\begin{abstract}
Full-body portrait stylization, which aims to translate portrait photography into a cartoon style, has drawn attention recently. However, most methods have focused only on converting face regions, restraining the feasibility of use in real-world applications. A recently proposed two-stage method expands the rendering area to full bodies, but the outputs are less plausible and fail to achieve quality robustness of non-face regions. Furthermore, they cannot reflect diverse skin tones. In this study, we propose a data-centric solution to build a production-level full-body portrait stylization system. Based on the two-stage scheme, we construct a novel and advanced dataset preparation paradigm that can effectively resolve the aforementioned problems. Experiments reveal that with our pipeline, high-quality portrait stylization can be achieved without additional losses or architectural changes.
\end{abstract}

\figTeaser{}



\maketitle

\section{Introduction}
The popularity of \textit{Webtoon}, a variety of comics published on digital platforms, has rapidly grown with the acceleration of comic digitization.
Recently, the industry has focused on utilizing the intellectual property of Webtoon to enhance  readers’ immersive experience. 
In particular, \textit{full-body portrait stylization}, which translates users’ portrait photos/videos into the desired Webtoon character, has been highlighted as a crucial technology.

Previous studies have tackled portrait stylization by translating face regions using StyleGAN2~\cite{karras2020analyzing} and shown a compelling visual quality~\cite{chong2021jojogan,kim2022cross}.
Nevertheless, adoption in real applications is still challenging because they constrain stylization to face regions. They also require “face alignment," which is detrimental to synthesizing arbitrary poses or non-faces.
To alleviate this, \citet{arcanegan} used a two-stage scheme in which they fine-tune a pre-trained StyleGAN2 with cartoon face images and then synthesized a large photo-cartoon paired dataset using the blending method~\cite{pinkney2020resolution}.
The paired dataset was used to train the image-to-image translation network, which served as the final stylization model (\Fref{fig:framework}a, c). With this, the translation network learns to stylize full-body portraits (including backgrounds) by referring to the facial textures.

Although such a strategy facilitates the stylization of non-face regions, it has several drawbacks that hinder its usage in real-world applications, as summarized in \Tref{table:contribution}. 
\textbf{(1)} The quality of face region is less appealing than that of previous face-only stylization methods~\cite{chong2021jojogan,kim2022cross}.
\textbf{(2)} The quality of non-face region is often unsatisfactory because the system manifests the color and texture of non-face regions with the distribution of faces and hair (from the training dataset) alone. 
\textbf{(3)} It tends to ignore the skin tone of users’ portraits and produces a \textit{procrustean} skin color (whitening mostly). 
This issue is serious, particularly when considering recent ethical issues regarding the racial biases of a deep generative model~\cite{Quach2020}.

In this study, we propose a data-centric approach toward practical full-body portrait stylization that is usable in real-world applications. 
Previous studies in this field have neglected the dataset perspective, although this often greatly impacts performance. 
Unfortunately, a high-quality Webtoon (or cartoon) dataset is not always accessible or it incurs a large cost to acquire. 
Instead, we focused on generating a synthetic training dataset using an advanced data preparation pipeline and augmentation methodology  (\Fref{fig:framework}b).

The proposed approach first separates source portrait images into head and background regions and then creates a synthetic dataset by stylizing each region individually (\Sref{subsec:synthesis}).
Unlike ours, previous methods jointly stylize both regions and train the stylization network using only facial images, thus the background region would become out-of-distribution. We alleviated this by modeling facial parts with a method specialized for generating a cartoon face~\cite{kim2022cross}, and the background part with the model trained to stylize non-facial region.
We also perform input-color correction to maintain the color tone of the input portrait, enabling the system to express racial diversity that most deep stylization methods cannot realize. 
This also promotes stylization quality because it aligns the distributions of the head and background regions (\Sref{subsec:reflection}).
We further augment images with the proposed \textit{CutFace} to increase the quality robustness (\Sref{subsec:augmentation}).
Subsequently, the portrait stylization network is trained with the paired dataset, which consists of the synthesized target domain and corresponding source domain images, in supervised manner (\Fref{fig:framework}c).
Since the data-centric procedures are used only in the training phase, no additional cost is required for inference. The experiments show that the proposed approach successfully produces a production-level stylization application without complex architectural designs or loss functions.

\tableContribution
Our contributions are as follows. 
\textbf{(1)} Our data-centric approach stylizes face, body parts, and background naturally with high quality using only a small number (<100) of Webtoon face images. 
\textbf{(2)} Our system reflects the skin color diversity that previous methods could not manage.
\textbf{(3)} Our system produces significantly fewer distortions that fulfill the robustness criteria of real-world applications.
\section{Approach}
The proposed solution follows two-stage stylization~\cite{arcanegan}, which manufactures a synthetic-paired dataset of source and target domain images (\Fref{fig:framework}a), then trains the stylization network in a supervised regime (\Fref{fig:framework}c).
Based on this, we aim on constructing a training dataset in elaborate manners by introducing data-centric procedures: full-body aware data synthesis, input color correction, and full-body aware data augmentation (\Fref{fig:framework}b). 
With these governing modules, we replace the first phase of the two-stage approach (\Fref{fig:framework}a) to enhance the target domain images (\textit{e.g.,} Webtoon or cartoon) without affecting the source domain images (\textit{e.g.,} FFHQ).
With our approach, we can stylize the input portrait naturally with high quality using only a small number (<100) of Webtoon face images.

\subsection{Full-Body Aware Data Synthesis}
\label{subsec:synthesis}
The two-stage method produces unsteady outputs when \textbf{(1)} face of the target character has severe geometric deformations or 
\textbf{(2)} the pixel distribution of background regions in the target domain images are diverged from that of the real source domain images.
For the former case, we observe that the blending-based method~\cite{pinkney2020resolution} is often vulnerable to the geometric changes of facial parts.
Consequently, the portrait stylization framework based on this method is also affected. Thus, we employ a cross-domain style-mixing (CDSM)~\cite{kim2022cross}, which is known to be robust to geometric deformation. With this, we generate source domain images $X_{src}$ through FFHQ pre-trained StyleGAN2~\cite{karras2020analyzing} and initial target domain images $X^{cdsm}_{tgt}$ via CDSM.

\figFramework

In our investigation, we observed that the quality degradation due to the latter scenario arises because the previous method trains the stylization network using only face images. Accordingly, the background regions received at the inference phase become out-of-distribution, resulting in an unexpected result. \Fref{fig:histogram}d and e show that the baseline (two-stage) method undesirably alters the shape of the original (FFHQ) pixel distribution. Consequently, it shifts the color and texture of the background while pursuing the distribution of head regions (e.g., purplish and flattened).

To resolve this, we generate target domain images by individually modeling head and background to separate their distributions. 
With source domain images $X_{src}$, we create corresponding head masks $M_{head}$, which include face, hair, and neck, with a pre-trained face parser~\cite{yu2021bisenet}. 
The masks are used to filter background regions of the CDSM output; \textit{i.e.,} $X^{cdsm}_{tgt} \odot M_{head}$.
We subsequently create synthetic background images $X^{sty}_{tgt}$ by translating $X_{src}$ via the background stylization model~\cite{chen2019animegan} then mask head regions; \textit{i.e.,} $X^{sty}_{tgt} \odot (1-M_{head})$.
Then, full-body aware images, $X^{fb}_{tgt}$ are formed by combining both regions.
$X^{fb}_{tgt}$ can be used when training the stylization network (\Fref{fig:framework}c) instead of images from the blending-based method, $X^{blend}_{tgt}$ that the vanilla two-stage approach utilized.
Note that the background stylization network is trained on diverse Webtoon images to translate the input to the universal Webtoon background style while adequately maintaining the input color (\Fref{fig:histogram}f). It also shows that our method ensures that the distribution is not markedly diverged from that of FFHQ (\Fref{fig:histogram}d).

As shown in \Fref{fig:ablation}, the proposed full-body-aware data synthesis (\Fref{fig:ablation}c) improves the stylization quality of the baseline (\Fref{fig:ablation}b). 
In particular, the background is rendered aesthetically, preserving the original color of source photo and expressing delicate and vibrant Webtoon styles. 
In contrast, the baseline not only fails to capture background content information in source photography but also shows notable artifacts on the face region.

\figHistogram

\subsection{Input Color Correction}
\label{subsec:reflection}
In this section, we perform input color correction to head regions.
The core motivation of this procedure is to improve the stylization quality of facial parts and increase the capability to handle racial diversity. 
The former inspiration regards relaxing the unaligned pixel distribution of head and background regions.
Generated images of full-body aware synthesis (\Sref{subsec:synthesis}),
$X^{fb}_{tgt}$, are composed of background regions, which are stylized while preserving
the input color (\Fref{fig:histogram}f) and head regions, which are stylized by referring to the color/texture of the target character (\Fref{fig:histogram}b).
That is, the color of the background should be retained while the counterpart for facial part should be substantially changed.
Such a complex mapping is difficult to learn without an explicit guide of the head position. Therefore, we relax the complication of this mapping function by simply changing to at least preserve noticeable color information of source image for head regions  (\textit{i.e.,} \Fref{fig:histogram}b $\rightarrow$ c).

The second motivation, racial diversity, arises to stylize while considering a person’s skin color, and not naively pursuing a character’s skin color. 
Ignoring skin tone is unpleasant for some users; thus building a stylization system that can achieve color diversity is critical.
As shown in \Fref{fig:histogram}b, without additional care with handling skin color, facial parts of the synthetic dataset tend to be white-washed when the target character has a light skin color owing to the data bias.
This problem is nontrivial; hence, in this work, we address this problem with a simple yet effective algorithm.

To reflect input color, we calculate the average color of facial parts of $X_{src}$
and $X^{cdsm}_{tgt}$, resulting in  $\bar{c}_{src}$ and $\bar{c}_{tgt}$, respectively. 
Then, we convert images and average colors to Lab color space and produce color reflected target images as: $X^{cr}_{tgt} = X^{cdsm}_{tgt} + (\bar{c}_{src} - \bar{c}_{tgt})$. 
The processed images are then converted back to RGB space.
With a simple color transfer algorithm, our method effectively reflects the color tone of the input portrait and can achieve color diversity  (\Fref{fig:ablation}g).
Now, the processed target domain images $X^{cr}_{tgt}$ attain the ability
to handle the geometric deformation of cartoon characters and preserve the color distribution of the source domain images, $X_{src}$.
Using a paired dataset of $(X_{src}, X^{cr}_{tgt})$, we train the stylization network to serve the final product at the inference phase (\Fref{fig:framework}c).

\subsection{Full-Body Aware Data Augmentation}
\label{subsec:augmentation}
In the portrait dataset, background regions occupy less portions (35.6\%) than head regions, and more importantly, most background regions are blurred; thus, complicated patterns present in the photograph do not appear. Thus, the stylization quality and robustness of background could be limited.
One possible approach involves adding some landscape images to the training dataset after stylizing them with the background stylization network utilized in \Sref{subsec:synthesis}.
However, naively including landscape images could harm the quality of face regions, as landscape images do not contain head regions; thus, the stylization network struggles to learn the implicit localization of head positions.

\figAblation
\figComp

To mitigate this, we propose \textit{CutFace}, a data augmentation to build a robust and high-quality full-body stylization framework.
It generates augmented images $\tilde{X}^{cr}_{tgt}$ by integrating stylized landscape images $X^{sty}_{bg}$ and color reflected face images $X^{cr}_{tgt}$ (\Fref{fig:framework}b).
In particular, $k$ face images are randomly pasted on the landscape images (without overlapping). 
Using \textit{CutFace}, we guide the stylization network to implicitly learn the position of head regions without explicit information from the user; the stylization quality is improved, and the robustness of the background region also increases significantly (\Fref{fig:ablation}d, g) producing fewer distortions for all cases.
\section{Experimental Results}
We compared our framework with AnimeGAN~\cite{chen2019animegan} and the two-stage method~\cite{arcanegan} using six Webtoon datasets. Because all the datasets have few images (<100), we trained AnimeGAN with a paired dataset generated using the blending-based method, which is identical to the two-stage scheme.
To quantitatively measure the stylization results, we conducted a user study. We asked 25 users which outcomes are the best for \textbf{(1)} quality of face regions, \textbf{(2)} robustness of background regions (\textit{i.e.,} existence of artifacts), and \textbf{(3)} correction of input color, such as skin color.
As shown in \Tref{table:comparison}, our approach outperforms the competitors by a large margin on all survey questions. 
The superiority of our solution is also shown in \Fref{fig:comp}; the proposed method is the only one that can generate diverse skin colors. 
It also produces the best quality on both the face and background, while adequately preserving the content of the input photography (\textit{e.g.,} hair).
\section{Conclusion}
In this study, we proposed a data-centric approach to build a practical full-body portrait stylization system. 
We constructed a sophisticated dataset creation pipeline based on real-world application requirements. 
This first synthesizes the target domain images in a full-body-oriented manner and then reflects the colors of the input image to the generated image. 
Finally, \textit{CutFace} is executed to boost the robustness of the system. Our method achieves production-level full-body portrait stylization under diverse circumstances.
\tableComp

\bibliographystyle{ACM-Reference-Format}
\bibliography{}

\appendix
\section{Additional Results}
\begin{figure}[ht]
\newcommand{\h}{20.6mm}
\newcommand{\himg}{0.01mm}
\newcommand{\hh}{27.7mm}
\newcommand{\hhimg}{0.01mm}
\centering
\makebox[\h][c]{\small(a) Source}\hspace{\himg}
\makebox[\h][c]{\small(b) Baseline}\hspace{\himg}
\makebox[\h][c]{\small(c) + FB-aware}\hspace{\himg}
\makebox[\h][c]{\small(d) + \textit{CutFace}}\hfill
\\
\includegraphics[width=\h]{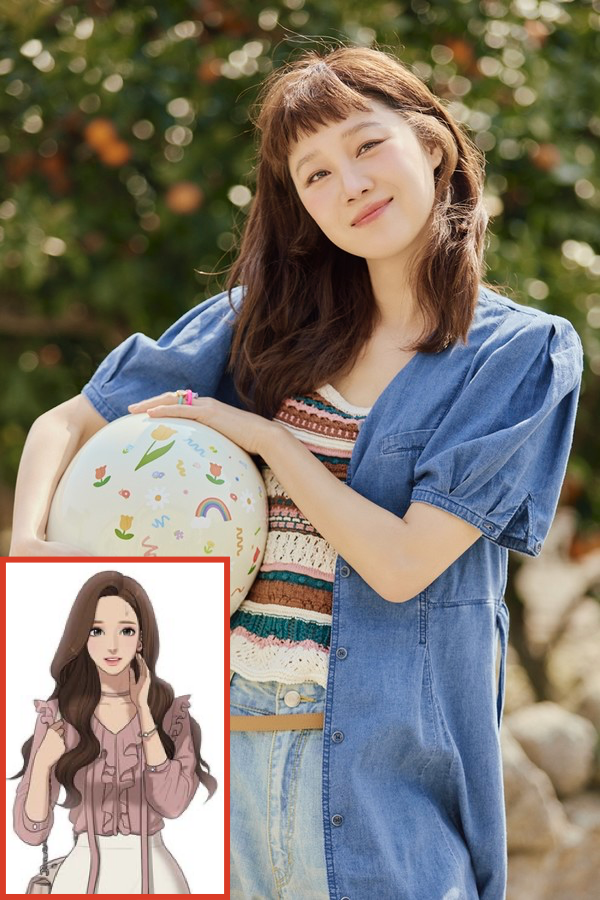}\hspace{\himg}
\includegraphics[width=\h]{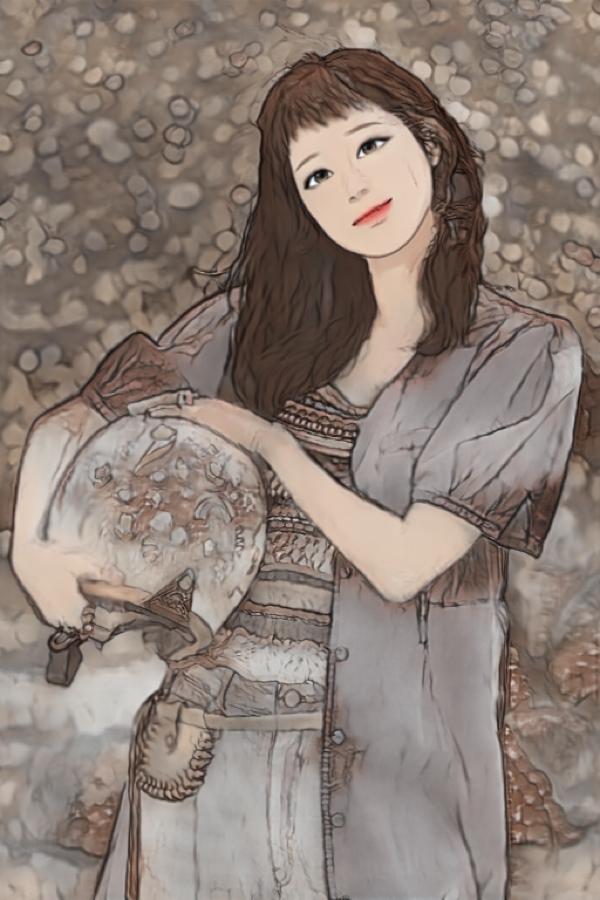}\hspace{\himg}
\includegraphics[width=\h]{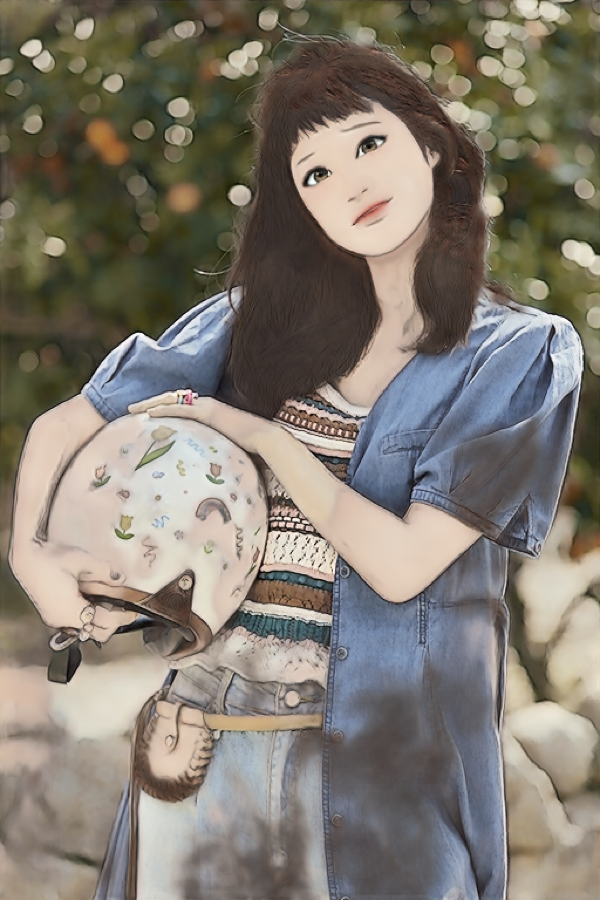}\hspace{\himg}
\includegraphics[width=\h]{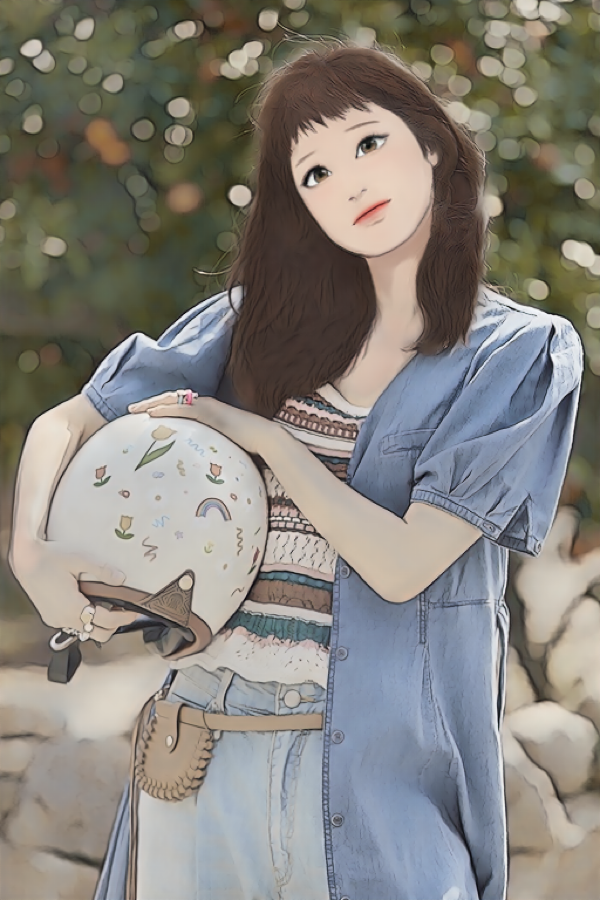}\hfill
\\
\vspace{-0.1em}
\makebox[\hh][c]{\small(e) Source}\hspace{\hhimg}
\makebox[\hh][c]{\small(f) w/o Color correction}\hspace{\hhimg}
\makebox[\hh][c]{\small(g) w/ Color correction}\hfill
\\
\includegraphics[width=\hh]{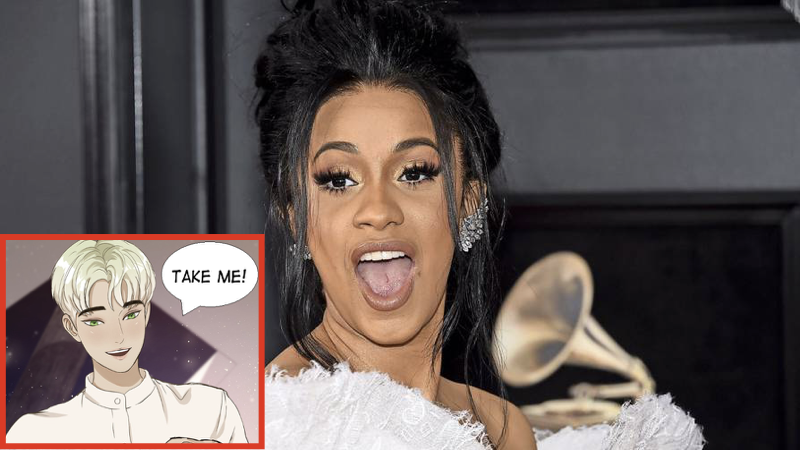}\hspace{\hhimg}
\includegraphics[width=\hh]{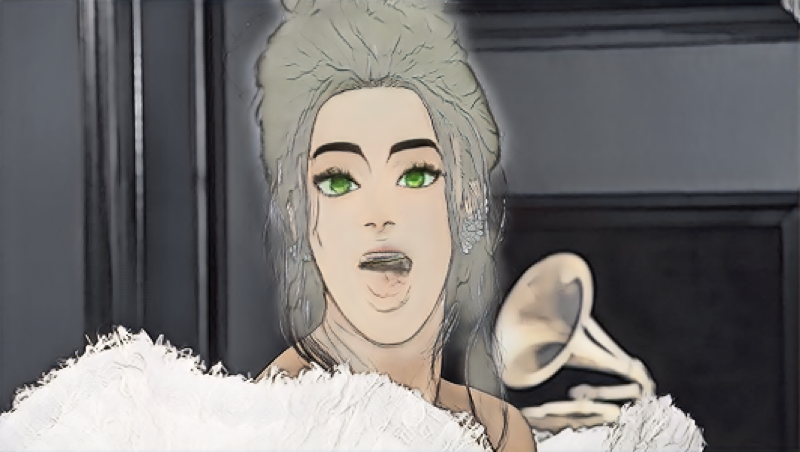}\hspace{\hhimg}
\includegraphics[width=\hh]{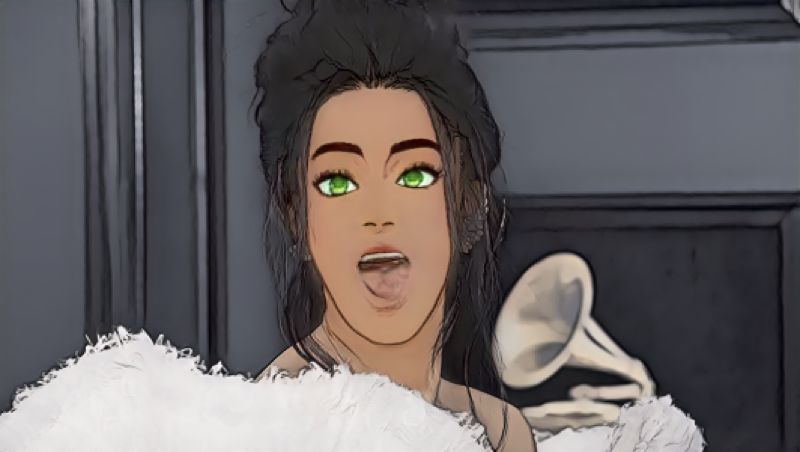}\hfill
\caption{\textbf{Ablation study.} Each proposed procedure successfully improves the quality, robustness, and ability to reflect skin color of the source photography. Baseline: two-stage approach~\cite{arcanegan}.}
\label{fig:appendix_ablation}
\end{figure}

\begin{figure}[ht]
\newcommand{\h}{41.0mm}
\newcommand{\himg}{0.01mm}
\centering
\makebox[\h][c]{\small(a) Source}\hspace{\himg}
\makebox[\h][c]{\small(b) AnimeGAN~\cite{chen2019animegan}}\hfill
\\
\includegraphics[width=\h]{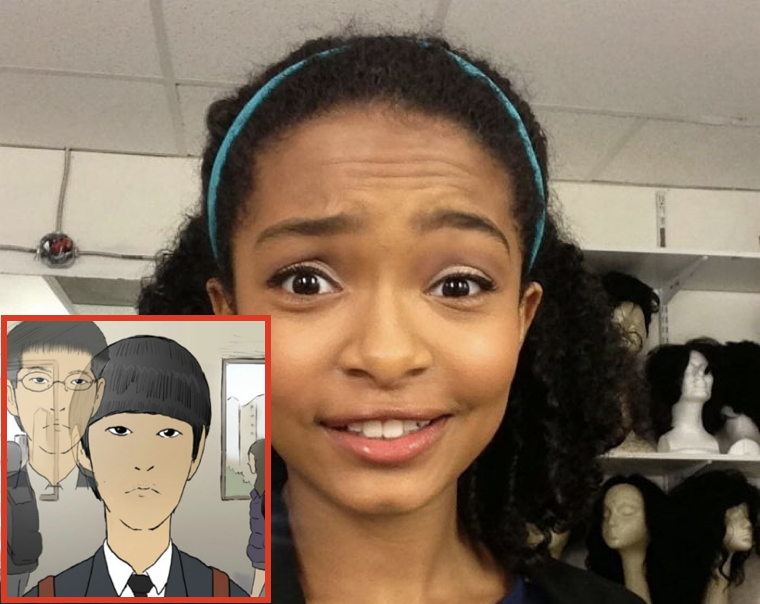}\hspace{\himg}
\includegraphics[width=\h]{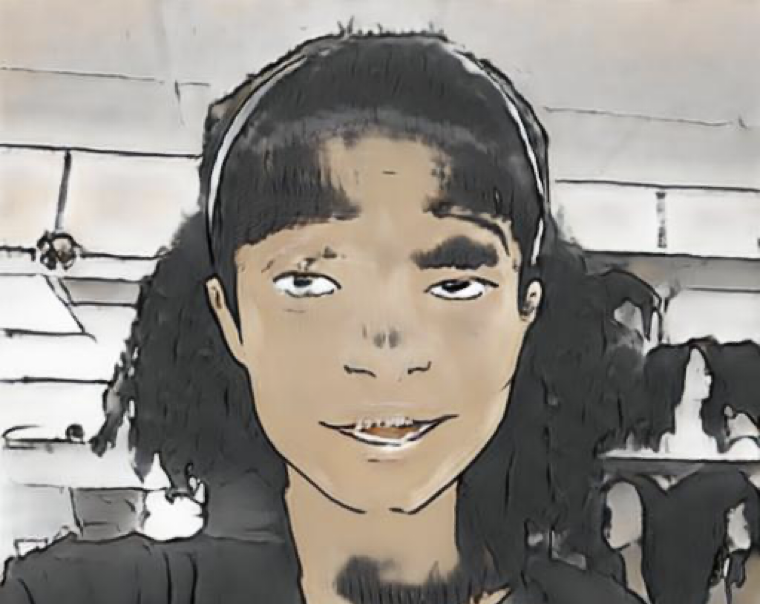}\hfill
\\
\makebox[\h][c]{\small(c) Two-stage~\cite{arcanegan}}\hspace{\himg}
\makebox[\h][c]{\small(d) WebtoonMe (ours)}\hfill
\\
\includegraphics[width=\h]{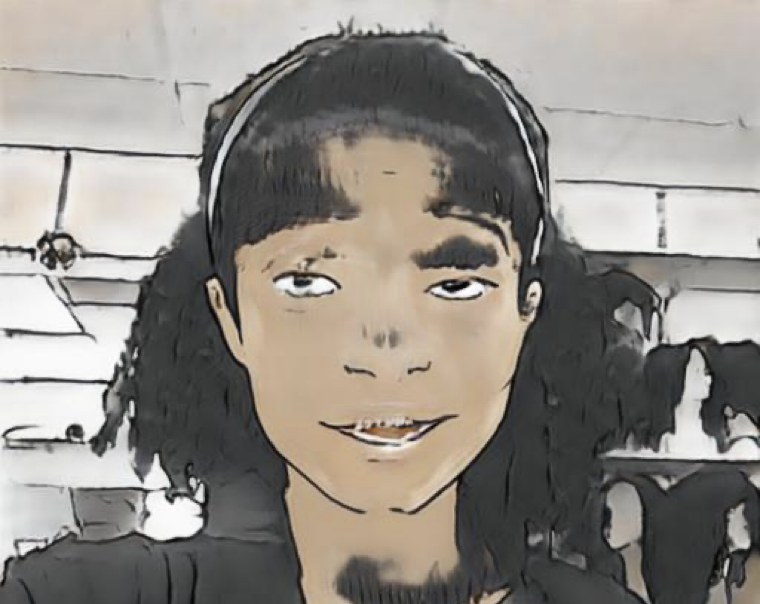}\hspace{\himg}
\includegraphics[width=\h]{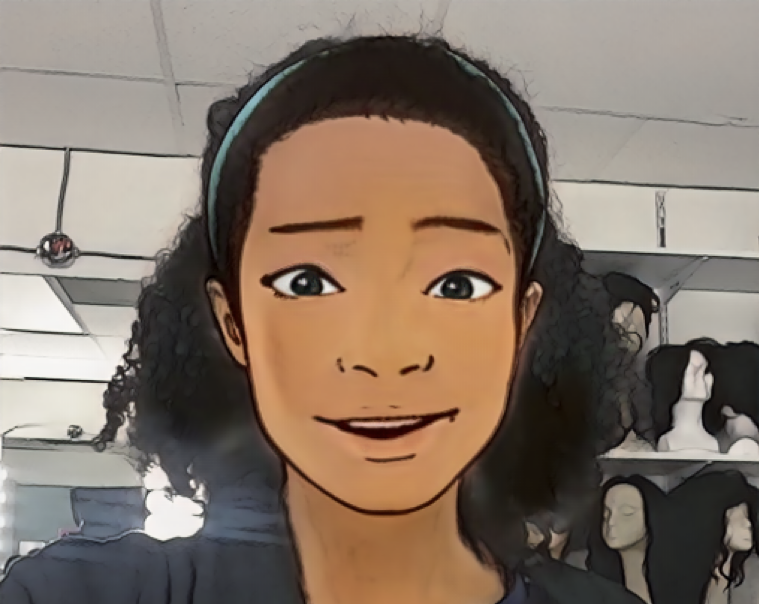}\hfill
\caption{\textbf{Qualitative comparison.} WebtoonMe shows superiority in both stylization quality and expression of color diversity.}
\label{fig:appendix_comp}
\end{figure}

\begin{figure*}[htp]
\centering
\newcommand{\h}{43mm}
\newcommand{\himg}{0.5mm}
\centering
\makebox[\h][c]{Input photography}\hspace{\himg}
\makebox[\h][c]{Stylization result}\hspace{1.0mm}
\makebox[\h][c]{Input photography}\hspace{\himg}
\makebox[\h][c]{Stylization result}\hfill
\\
\includegraphics[width=\h]{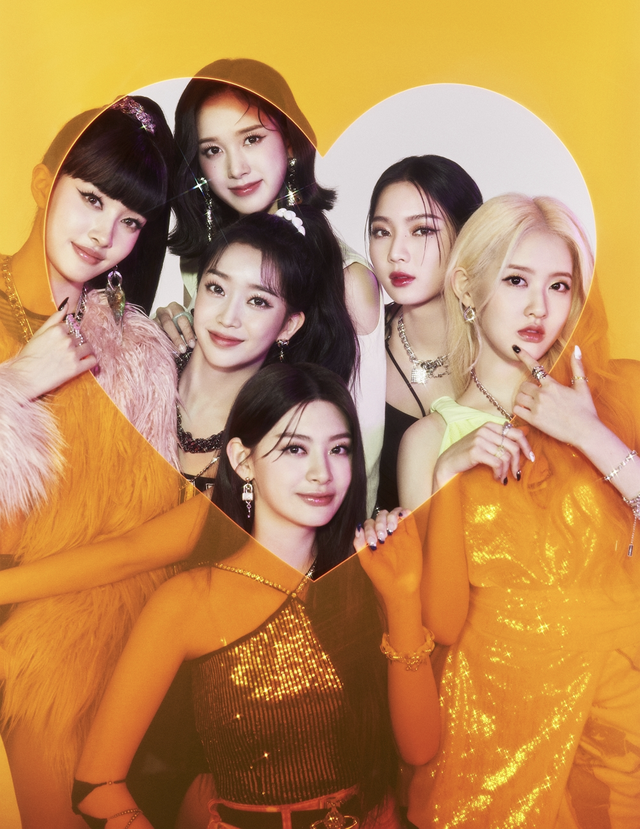}\hspace{\himg}
\includegraphics[width=\h]{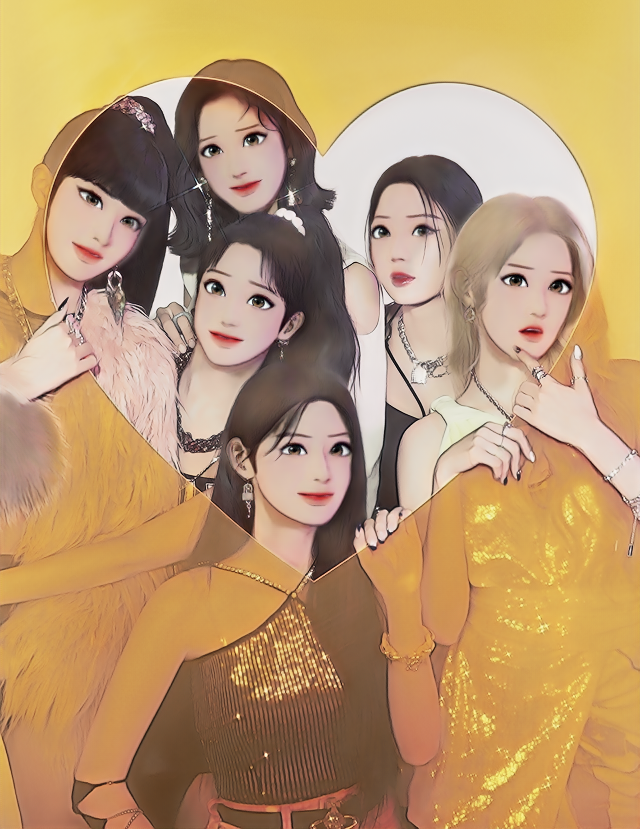}\hspace{1.0mm}
\includegraphics[width=\h]{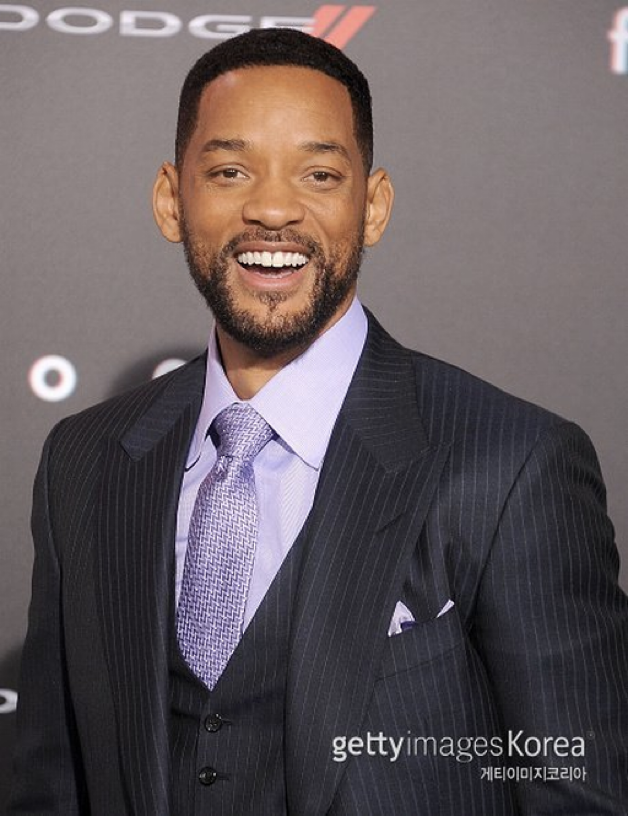}\hspace{1.0mm}
\includegraphics[width=\h]{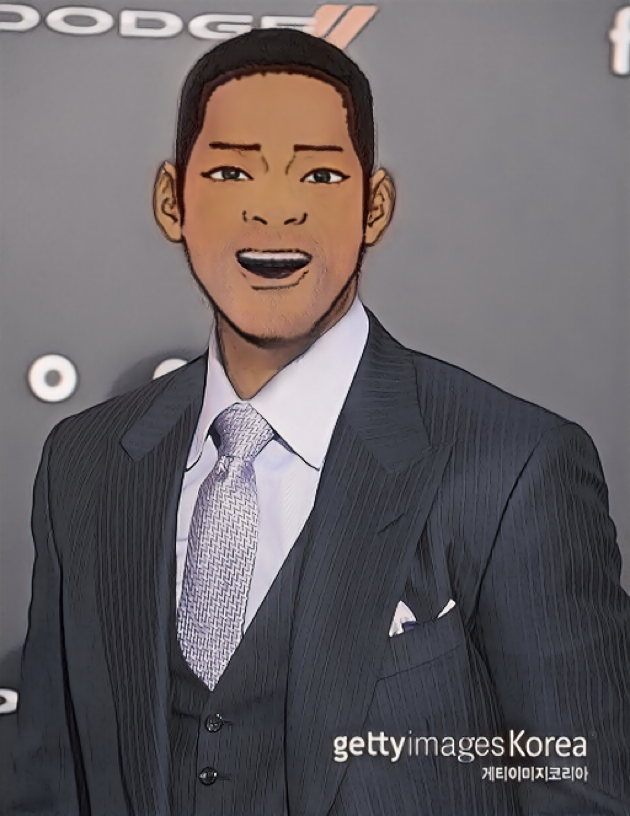}\hspace{\himg}\hfill
\\\vspace{1mm}
\includegraphics[width=\h]{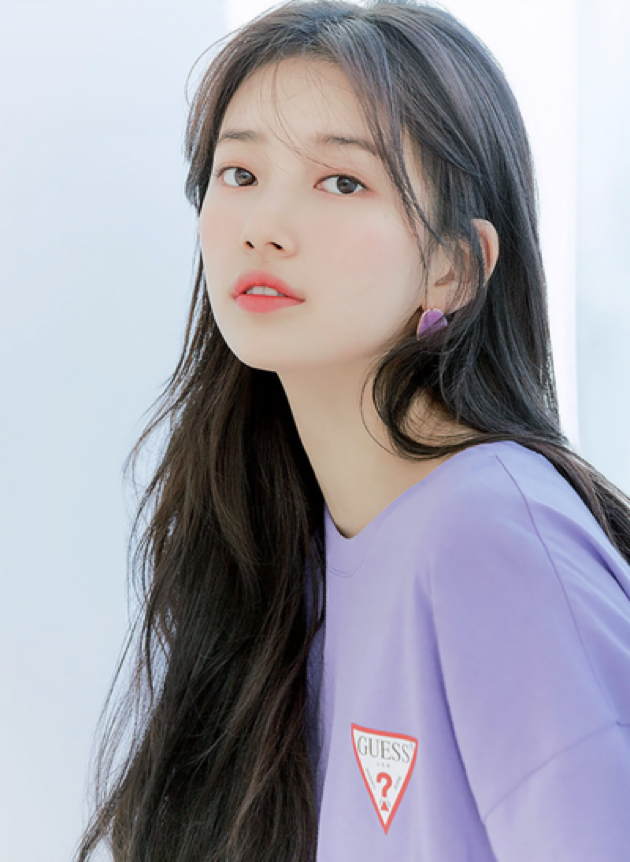}\hspace{\himg}
\includegraphics[width=\h]{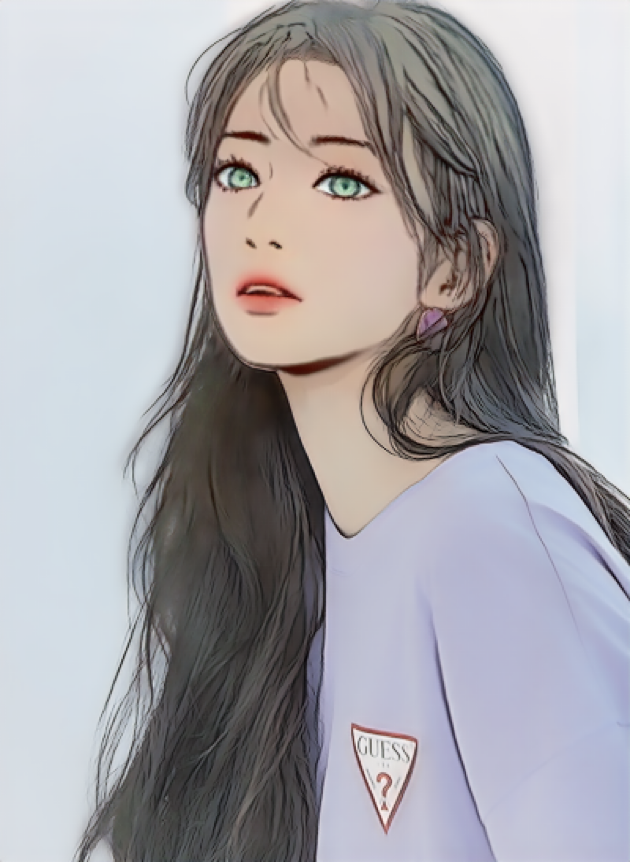}\hspace{1.0mm}
\includegraphics[width=\h]{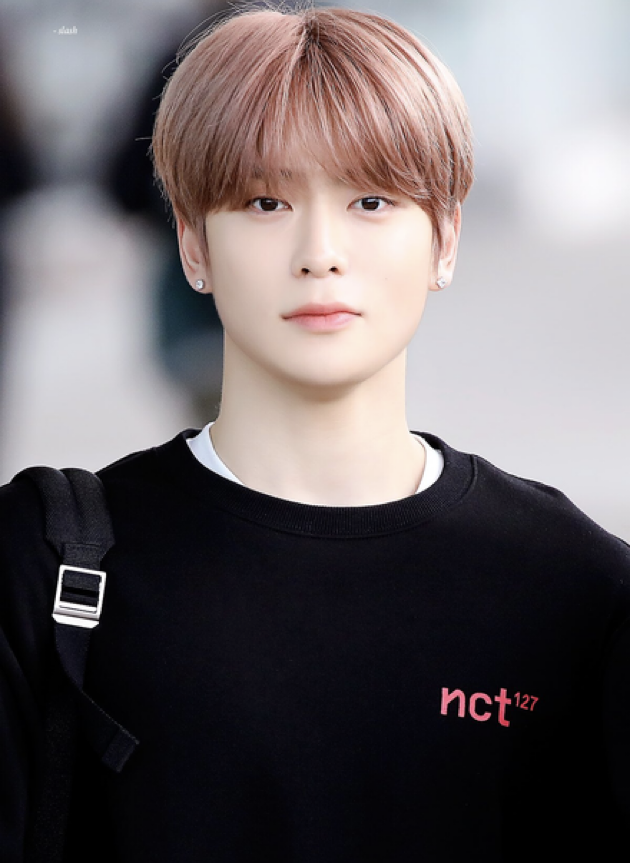}\hspace{1.0mm}
\includegraphics[width=\h]{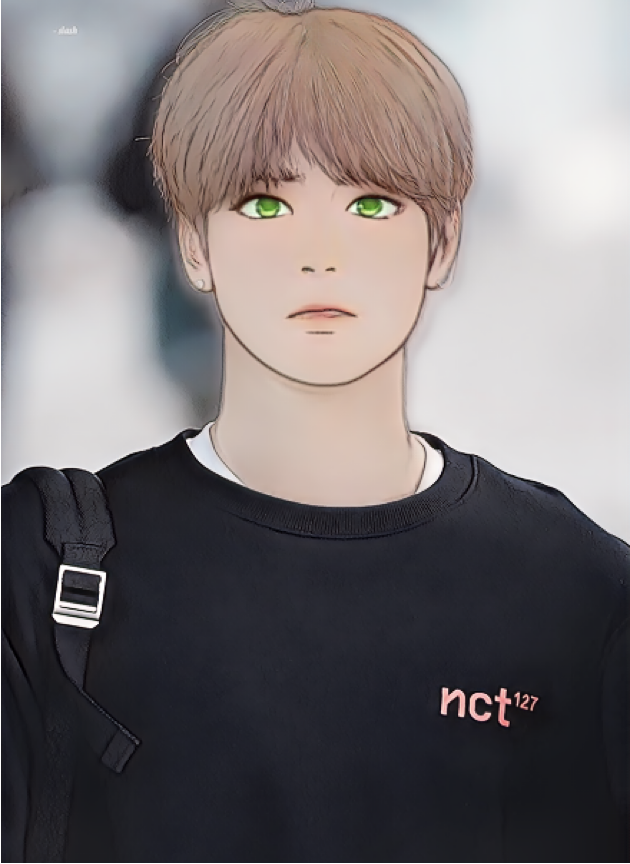}\hspace{\himg}\hfill
\caption{\textbf{Additional Stylization results.}}
\label{fig:appendix1}
\end{figure*}

\begin{figure*}[htp]
\centering
\newcommand{\h}{87mm}
\newcommand{\himg}{0.5mm}
\centering
\makebox[\h][c]{Input photography}\hspace{\himg}
\makebox[\h][c]{Stylization result}\hfill
\\
\includegraphics[width=\h]{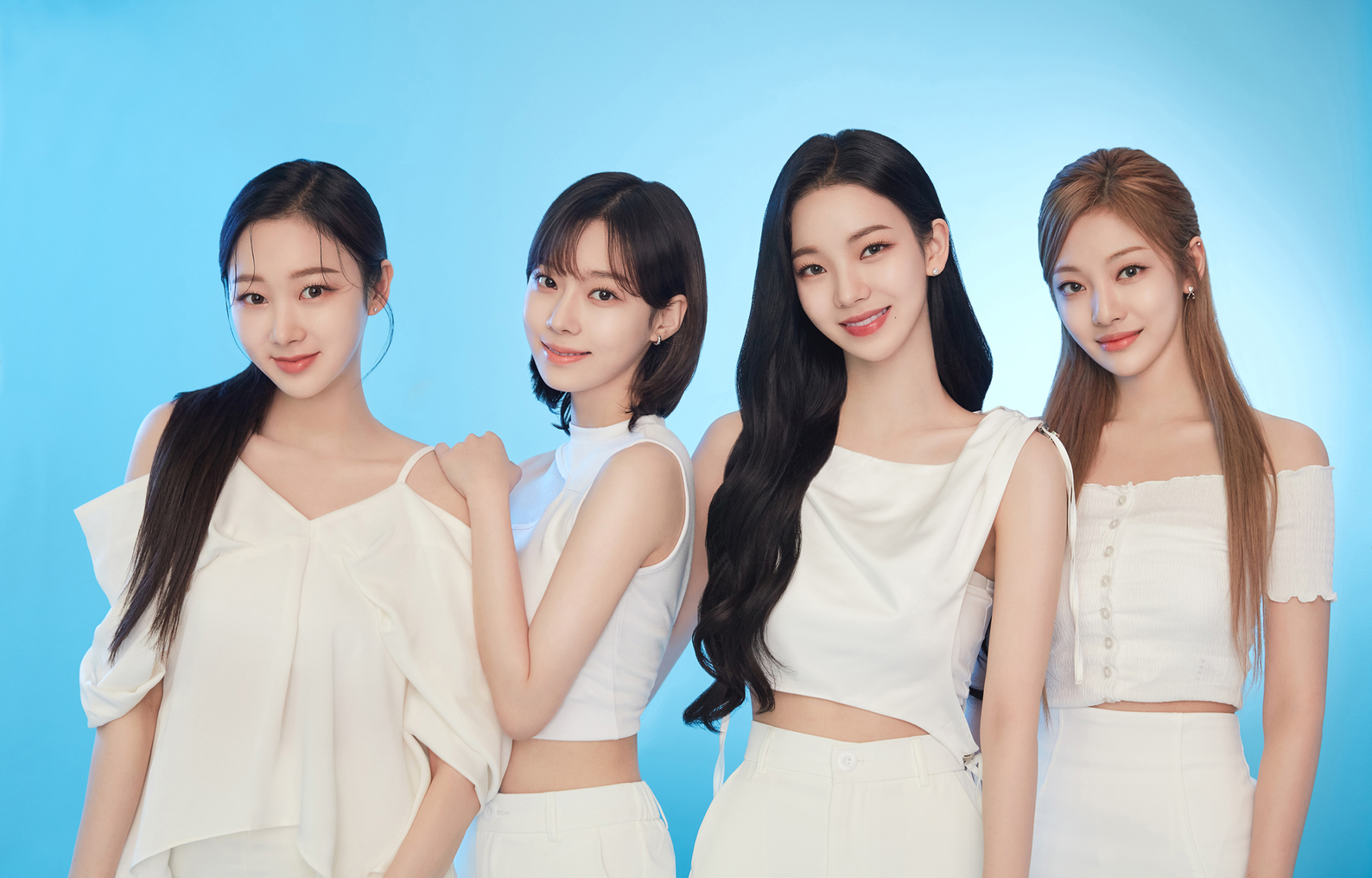}\hspace{\himg}
\includegraphics[width=\h]{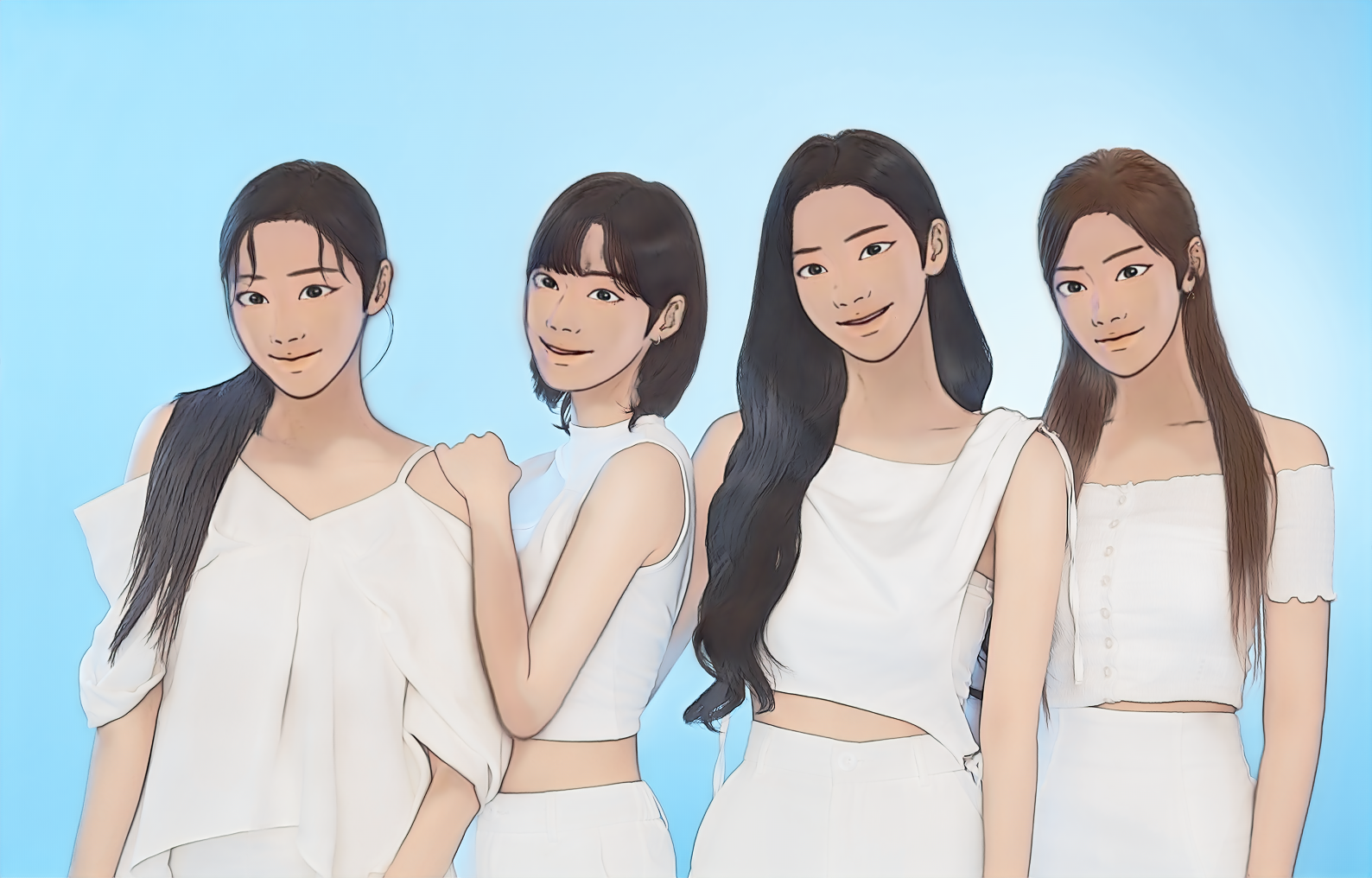}\hfill
\\\vspace{1mm}
\includegraphics[width=\h]{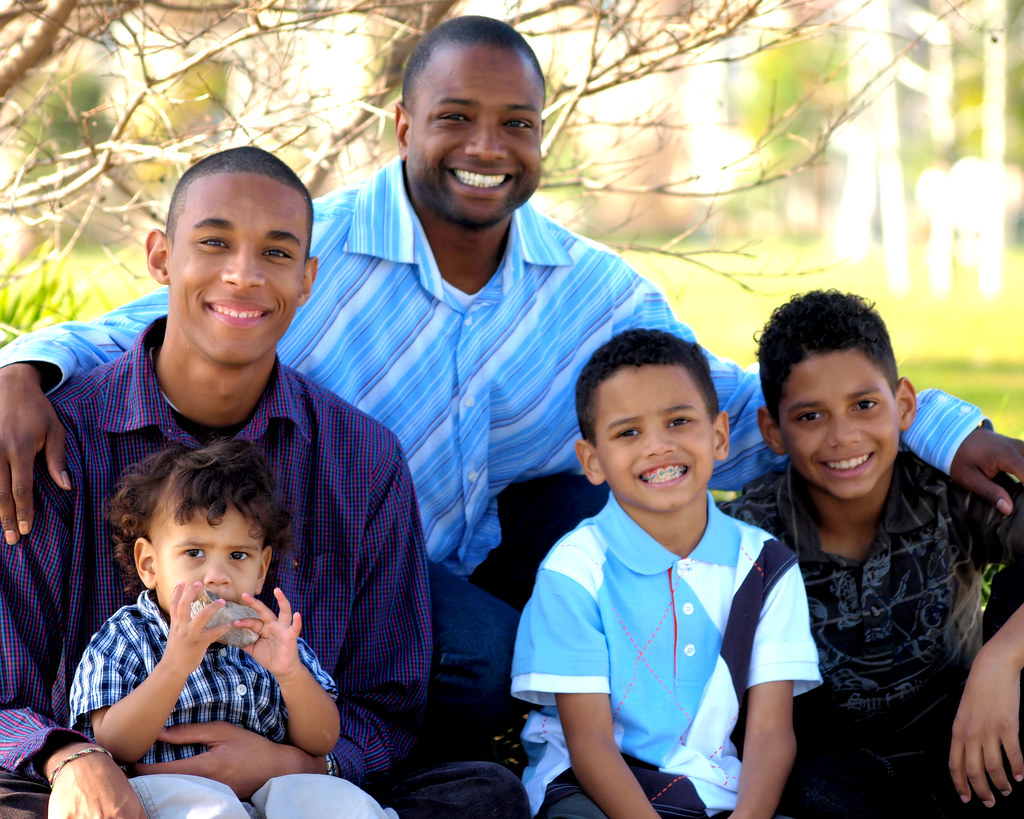}\hspace{\himg}
\includegraphics[width=\h]{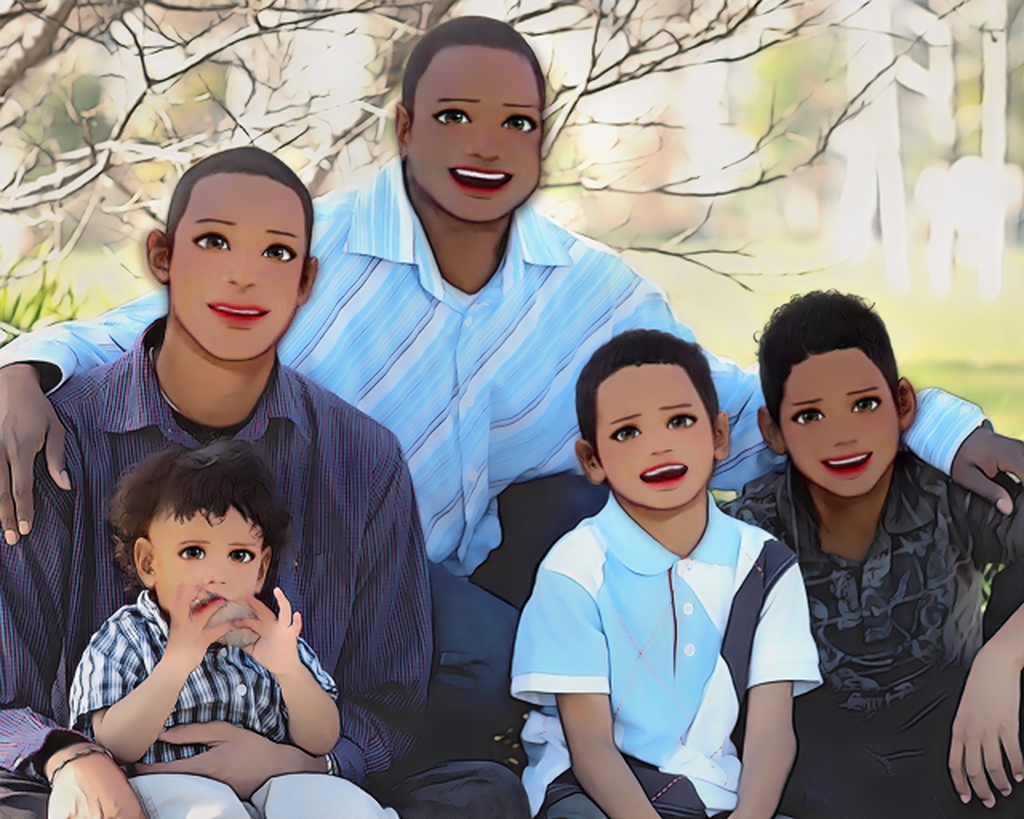}\hfill
\\\vspace{1mm}
\includegraphics[width=\h]{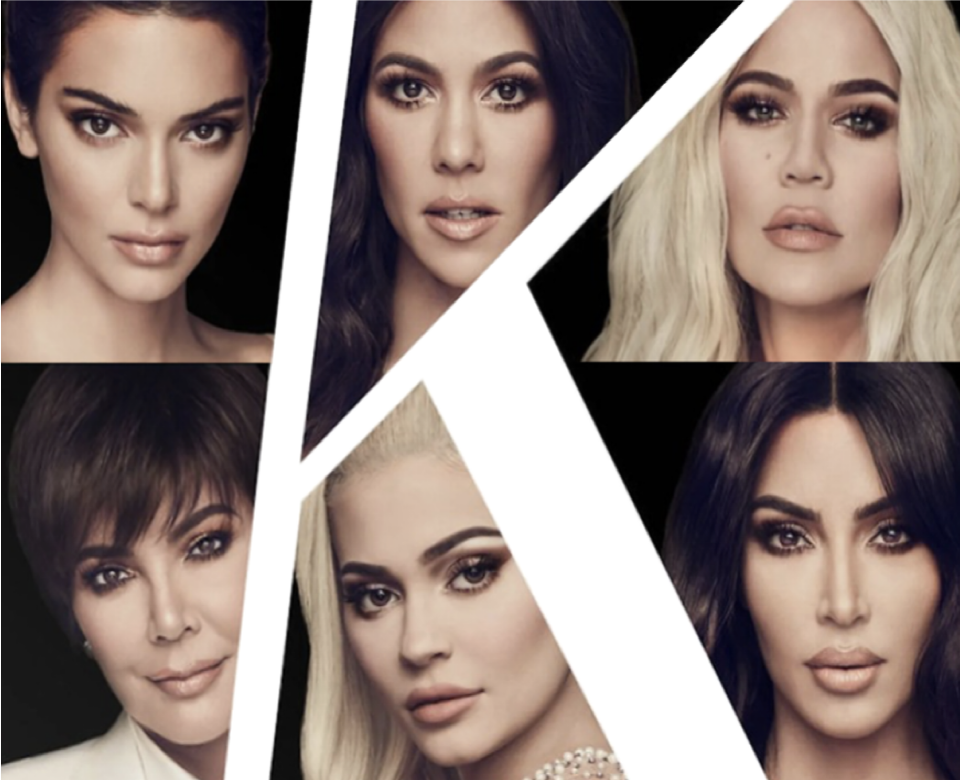}\hspace{\himg}
\includegraphics[width=\h]{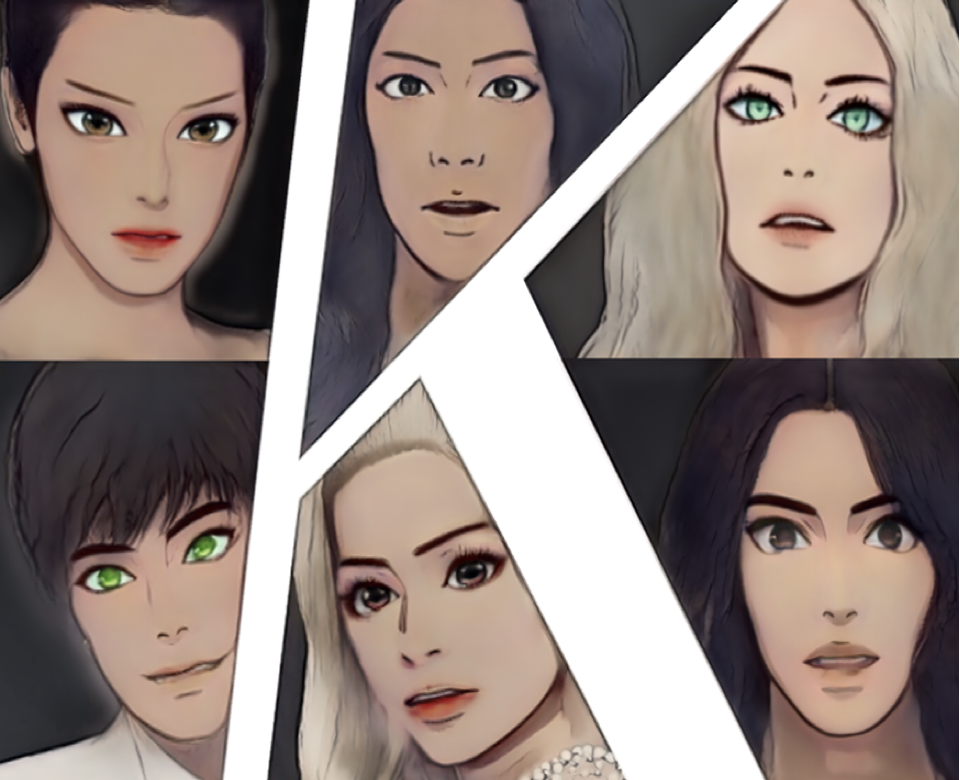}\hfill
\caption{\textbf{Additional Stylization results.}}
\label{fig:appendix2}
\end{figure*}

\begin{figure*}[htp]
\centering
\newcommand{\h}{87mm}
\newcommand{\himg}{0.5mm}
\centering
\makebox[\h][c]{Input photography}\hspace{\himg}
\makebox[\h][c]{Stylization result}\hfill
\\
\includegraphics[width=\h]{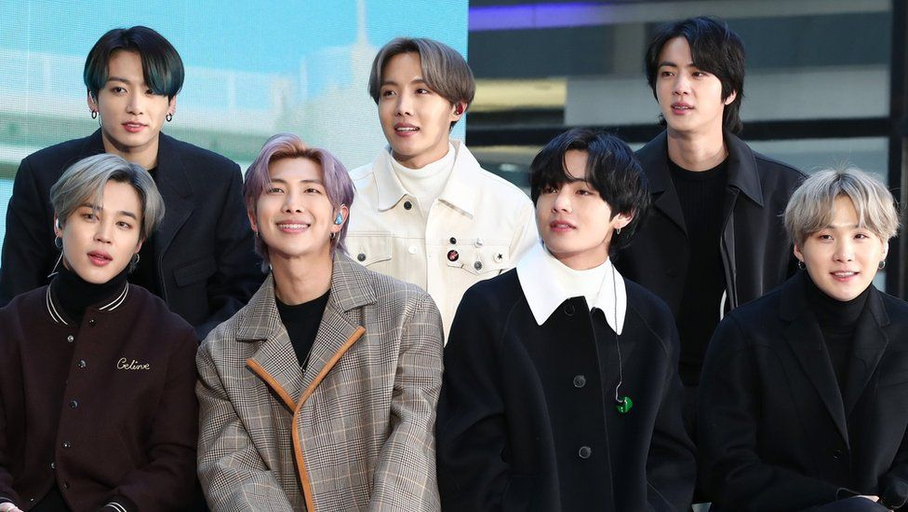}\hspace{\himg}
\includegraphics[width=\h]{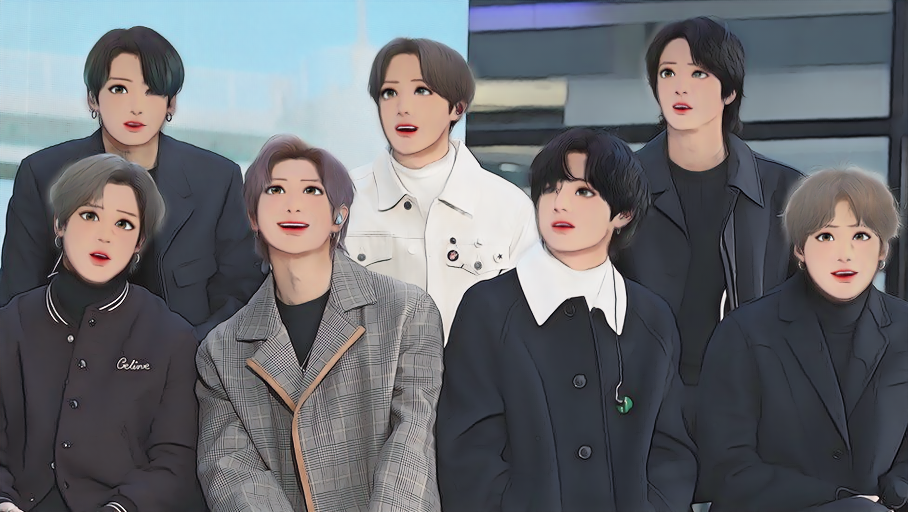}\hfill
\\\vspace{1mm}
\includegraphics[width=\h]{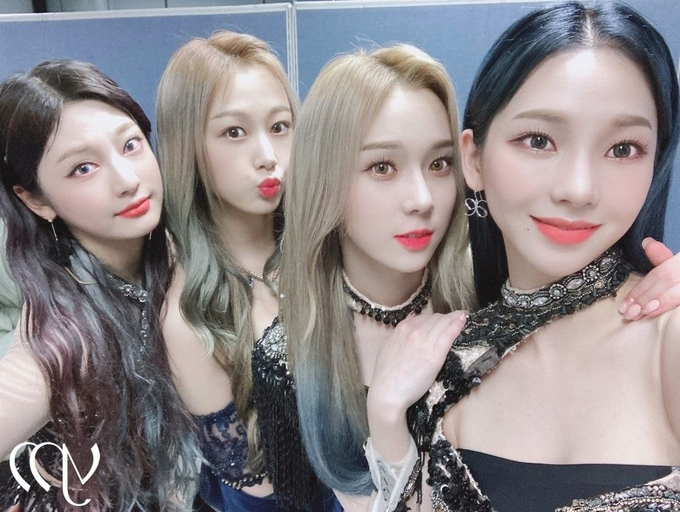}\hspace{\himg}
\includegraphics[width=\h]{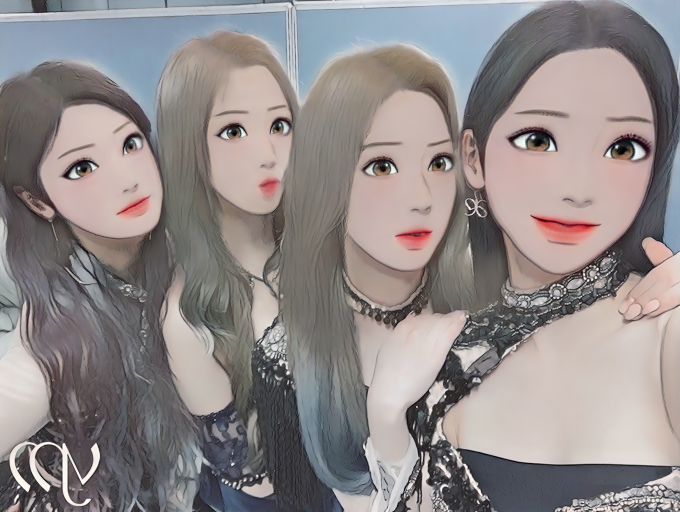}\hfill
\caption{\textbf{Additional Stylization results.}}
\label{fig:appendix3}
\end{figure*}

\Fref{fig:appendix_ablation} and \ref{fig:appendix_comp} show additional results of ablation study and qualitative comparison.
\Fref{fig:appendix1}, \ref{fig:appendix2}, and \ref{fig:appendix3} present more stylization results.

\end{document}